\pdfoutput=1

\ifdefined\REVIEW
    \documentclass[
        a4paper,
    ]{article}
    \usepackage{lineno, setspace}
    \modulolinenumbers[1]
    \pagewiselinenumbers
\else
    \documentclass[
        a4paper,
        twocolumn
    ]{article}
\fi

\usepackage{IEK10} \usepackage{natbib}

\def\IEK10{
  Forschungszentrum Jülich GmbH,
  Institute of Climate and Energy Systems,
  Energy Systems Engineering (ICE-1),
  Jülich 52425,
  Germany
}
\def\SVT{
  RWTH Aachen University,
  Process Systems Engineering (AVT.SVT),
  Aachen 52074,
  Germany
}
\def\JARA{
  JARA-ENERGY,
  Jülich 52425,
  Germany
}
\def\RWTH{
  RWTH Aachen University,
  Aachen 52062,
  Germany
}

\newcommand{\mytitle}{
End-to-End Reinforcement Learning of Koopman Models for Economic Nonlinear Model Predictive Control
}

\newcommand{\affil}{
  \begin{itemize}[leftmargin=3mm, itemsep=0mm]
    \item[$^a$]\IEK10
    \item[$^b$]\SVT
    \item[$^c$]\JARA
    \item[$^d$]\RWTH
  \end{itemize}
}

\def\firstAuthor{Daniel Mayfrank}
\newcommand{\myauthor}{
\firstAuthor$^{a,d}$\orcidlink{0009-0000-6275-0614}, 
Alexander Mitsos$^{c,a,b}$\orcidlink{0000-0003-0335-6566}, 
Manuel Dahmen$^{a,*}$\orcidlink{0000-0003-2757-5253} }
\newcommand{\correspondingauthor}{Manuel Dahmen, \IEK10 \newline E-mail: m.dahmen@fz-juelich.de}
\author{\myauthor}

\usepackage[
  colorlinks,
  linkcolor=blue,
  citecolor=blue,
  urlcolor=blue,
  pdftitle={\mytitle},
  pdfauthor={\firstAuthor}
]{hyperref}
\usepackage[capitalise, nameinlink]{cleveref}
\crefname{table}{Tab.}{Tab.}

\usepackage{orcidlink}

\newcommand{\setpgfexternalcounter}[1]{
  \makeatletter \pgfkeysgetvalue{/tikz/external/figure name}\myexternalname
  \expandafter\gdef\csname c@tikzext@no@\myexternalname\endcsname{#1}\makeatother
}

\begin{document}

\ifx\REVIEW\undefined
\twocolumn[
\begin{@twocolumnfalse}
\fi
  \thispagestyle{firststyle}

  \begin{center}
    \begin{large}
      \textbf{\mytitle}
    \end{large} \\
    \myauthor
  \end{center}

  \vspace{0.5cm}

  \begin{footnotesize}
    \affil
  \end{footnotesize}

  \vspace{0.5cm}

    (Economic) nonlinear model predictive control ((e)NMPC) requires dynamic models that are sufficiently accurate and computationally tractable.
    Data-driven surrogate models for mechanistic models can reduce the computational burden of (e)NMPC; however, such models are typically trained by system identification for maximum prediction accuracy on simulation samples and perform suboptimally in (e)NMPC.
    We present a method for end-to-end reinforcement learning of Koopman surrogate models for optimal performance as part of (e)NMPC.
    We apply our method to two applications derived from an established nonlinear continuous stirred-tank reactor model. The controller performance is compared to that of (e)NMPCs utilizing models trained using system identification, and model-free neural network controllers trained using reinforcement learning.
    We show that the end-to-end trained models outperform those trained using system identification in (e)NMPC, and that, in contrast to the neural network controllers, the (e)NMPC controllers can react to changes in the control setting without retraining.

  \vspace*{5mm}
\ifx\REVIEW\undefined
\end{@twocolumnfalse}
]
\fi

\newpage

\section{Introduction}\label{sec:intro}
Data-driven surrogate models present a promising avenue for rendering economic nonlinear model predictive control (eNMPC) tractable by reducting the computational demand associated with solving the underlying optimal control problem (\cite{mcbride2019overview}). System identification (SI) represents the predominant methodology employed for training data-driven dynamic surrogate models; however, SI narrowly focuses on optimizing average prediction accuracy on a collection of simulation samples. In contrast, recent studies have demonstrated the superiority of dynamic surrogate models, trained explicitly for optimal performance in control applications via reinforcement learning (RL) techniques, over SI-trained models (\cite{amos2018differentiable, yin2022embedding, iwata2022data, chen2019gnu, gros2019data}). Nonetheless, these findings were limited to the learning of linear models (\cite{chen2019gnu}), applications devoid of state constraints (\cite{amos2018differentiable, yin2022embedding, iwata2022data, chen2019gnu}), or primarily concentrated on the adaptation of bounds and cost function rather than the dynamic MPC model itself (\cite{gros2019data}).

RL research primarily focuses on learning model-free control policies that eschew the utilization of system state predictions to dictate control actions. In contrast, learning a dynamic model and employing its predictions to derive a control law, as exemplified in eNMPC, has crucial advantages: (i) retraining is rendered unnecessary should constraints or objective functions be altered, provided that the system dynamics persist unaltered; (ii) learning the system dynamics may be more sample-efficient than learning a model-free control policy (\cite{amos2018differentiable, chen2019gnu}); (iii) MPC has a well-established theory regarding performance and stability guarantees, especially for linear models and recent publications aim to extend this theory to include (learned) eNMPC (\cite{gros2019data, angeli2011average}).

Using RL to train dynamic surrogate models promises to combine the aforementioned advantages of model-based policies with the typical advantage of end-to-end learning over SI, i.e., the task-optimal performance of the resulting model. To train dynamic surrogate models using RL algorithms, (eN)MPC is viewed as a policy whose control outputs can be differentiated with respect to the parameters of the surrogate model. This requires differentiating a solution point of a parametric optimization problem with respect to the parameters (\cite{fiacco1985sensitivity, fiacco1990sensitivity, ralph1995directional}). \cite{lewis2009reinforcement} and \cite{lewis2012reinforcement} show that RL can be used to learn goal-directed dynamic models by connecting RL to the linear quadratic regulator (\cite{brunton2022data}), a model predictive feedback controller for control settings with linear dynamics and quadratic cost functions. \cite{amos2018differentiable} and \cite{gros2019data} study how to learn MPC controllers end-to-end by viewing these as differentiable policies. \cite{amos2018differentiable} use imitation learning (\cite{hussein2017imitation}), a supervised learning approach aiming to train a control policy through expert demonstrations, instead of RL to learn an NMPC but do not consider control settings with inequality constraints on state variables. Moreover, they learn a parameterized version of the true underlying dynamic equations, which requires an a priori known structure of the system dynamics. \cite{gros2019data} use RL in the form of deep Q-learning, a popular RL algorithm for environments with discrete action spaces and the first RL algorithm to claim superhuman performance on an Atari game (\cite{mnih2013playing}), to learn linear MPC and eNMPC controllers. However, in the nonlinear case, they focus on learning advantageous parameterizations of the constraints and cost function and only parameterize the dynamic model with a learnable constant added to the transition function. \cite{chen2019gnu} use both imitation learning and RL together with the differentiable MPC solver by \cite{amos2018differentiable} to learn task-optimal linear dynamic models for HVAC control without constraints on state variables.

Given assumptions we describe in Subsection \ref{sec:method_POSA}, a solution point $\bm{a}^{*}$ of a parametric optimization problem is differentiable with respect to the parameters $\bm{\theta}$ locally at a given $\bm{\theta}^{*}$. One of the assumptions is that the second-order sufficient conditions (SOSC) for a local minimum hold at $\bm{a}^{*}$, which, for convex parametric programs with strictly convex objective functions, is easily verifiable through the Karush-Kuhn-Tucker (KKT) conditions (\cite{karush1939minima, boyd2004convex}). Therefore, for end-to-end learning of dynamic models for MPC, it is desirable to learn dynamic models that lead to convex optimal control problems (OCPs). A simple condition for convex OCPs are formulations with linear dynamics and convex constraints. Koopman theory (\cite{koopman1931hamiltonian}) aims to find globally-valid linear approximations of nonlinear process dynamics through a nonlinear coordinate transformation into a higher-dimensional linear state space. Koopman theory and its extension to actuated systems for control applications (\cite{korda2018linear}, therefore, provides a powerful framework to obtain surrogate models that can accurately represent continuous nonlinear dynamical systems while still resulting in convex OCPs, especially when combined with modern deep learning algorithms (\cite{lusch2018deep}). \cite{iwata2022data} and \cite{yin2022embedding} both use RL for end-to-end learning of task-optimal Koopman models to obtain gain matrices that act on the Koopman invariant subspaces, minimizing the squared distance of the system state from a desired target state. However, similarly to linear-quadratic regulator control (\cite{kalman1960contributions}), this approach cannot easily be extended to (e)NMPC applications or applications with hard bounds on state variables.

Koopman-based MPCs have seen application in various domains due to their favorable balance between high representational capacity and comparatively low computational demand in solving the resulting OCPs. These applications range from ground vehicle (\cite{Cibulka20204}) to air vehicle control (\cite{Folkestad2021}), flow control (\cite{arbabi2018datadrivenkoopman}), and soft robotics (\cite{bruder2019modelingcontrolsoftrobots}). Some contributions have also studied the application of Koopman-based MPCs to control of chemical processes. \cite{Nasasingam2019} integrate Koopman models with Lyapunov-based MPC, thus guaranteeing controller feasibility and closed-loop stability. \cite{Nasasingam2020} integrate a priori system knowledge in the choice of the Koopman basis functions and achieve successful data-driven control of a hydraulic fracturing process. Using a learned disturbance estimator, \cite{SonKwonJProc2022} develop offset-free Koopman Lyapunov-based MPC to compensate for the unavoidable plant-model-mismatch resulting from a finite-dimensional approximation of the Koopman operator. They showcase their approach on the control of a batch pulp digester (\cite{SonKwon2021}). For hybrid systems with differing local dynamics, \cite{SonKwon2022} develop a hybrid Koopman MPC approach by clustering the training data and by training one local Koopman model for each cluster. Furthermore, \cite{Albalawi2023EconomicKoopman} demonstrate the efficacy of Koopman-based economic MPC on a continuous stirred-tank reactor (CSTR) case study.

We present a method for end-to-end RL of Koopman models for optimal performance in (e)NMPC applications with hard constraints on states. Using the \textit{cvxpylayers} package (\cite{Agrawal2019differentiable}) to construct differentiable Koopman-MPC policies, we use Proximal Policy Optimization (PPO) (\cite{schulman2017proximal}), a prominent RL algorithm for tasks with continuous action spaces, to learn task-optimal Koopman models. We test our approach on two distinct case studies derived from a continuous stirred-tank reactor model from the literature (\cite{flores2006simultaneous, du2015time}): (i) an NMPC case study, wherein the controller objective is to stabilize the product concentration in the presence of an externally varying product flow rate, and (ii) a demand response case study, in which eNMPC seeks to minimize electricity costs without violating constraints on state variables. We assess the resulting control performance by comparing it to that of dynamic models trained exclusively with SI as well as model-free policies trained by RL.

Our findings demonstrate that end-to-end learning consistently outperforms SI in learning dynamic surrogate models of the CSTR model in our control applications. Additionally, we investigate to which extent the learned models and policies still function in the case of modified constraints. Such a change in the control setting is an example of so-called \textit{distribution shift} (\cite{zhang2021adaptive}), which is a major challenge in machine learning applications (\cite{quinonero2008dataset, zhang2021adaptive, yao2022wild}). We show that, unlike model-free policies, MPCs employing end-to-end learned dynamic surrogate models may successfully adapt to such a change in the control setting without a need for re-training. To our knowledge, this work is the first to connect end-to-end learning of Koopman models to control applications with constraints on system variables and the first to study the ability of end-to-end learned dynamic models to adapt to a distribution shift in control.

The remainder of this paper is organized as follows: Section \ref{sec:method} provides the theoretical background to our work and presents our method. Section \ref{sec:results} provides numerical examples on a case study. Section \ref{sec:conclusion} discusses the conclusions and directions for future work.

\section{Method}\label{sec:method}
Subsections \ref{sec:method_Koopman4control} - \ref{sec:method_POSA} provide the necessary theory upon which we build our approach for end-to-end RL of dynamic Koopman surrogate models for MPC applications (Subsection \ref{sec:method_e2eKoopman}).

\subsection{Koopman theory for control}\label{sec:method_Koopman4control}
    The objective of applied Koopman theory is to obtain globally valid linear representations of nonlinear dynamics by lifting the system states into a higher dimensional Koopman space through a nonlinear coordinate transformation (\cite{koopman1931hamiltonian}). Multiple ways to extend Koopman theory to controlled systems have been proposed (e.g., \cite{proctor2018generalizing, williams2016extending, korda2018linear}). We use the approach by \cite{korda2018linear} as it results in convex OCPs, under the easily verifiable condition  that the objective function and inequality constraint functions are convex. The resulting models are of the form
    \begin{subequations}\label{eq:Koopman_model}
    	\begin{align}
            \bm{z}_0 &= \bm{\psi}_{\bm{\theta}}(\bm{x}_0), \label{eq:Koopman_model_encoder}\\
            \bm{z}_{t+1} &= \bm{A}_{\bm{\theta}} \bm{z}_t + \bm{B}_{\bm{\theta}} \bm{u}_t, \label{eq:Koopman_model_predictor}\\
            \hat{\bm{x}}_t &= \bm{C}_{\bm{\theta}} \bm{z}_t, \label{eq:Koopman_model_decoder}
    	\end{align}
    \end{subequations}
    where $\bm{z}_t \in \mathbb{R}^N$ is the lifted latent space vector, $\hat{\bm{x}}_t$ is the prediction of the states $\bm{x}_t \in \mathbb{R}^n$, and $\bm{u}_t \in \mathbb{R}^m$ is the control input vector. $\bm{A}_{\bm{\theta}} \in \mathbb{R}^{N \times N}$ and $\bm{B}_{\bm{\theta}} \in \mathbb{R}^{N \times m}$ together form an approximation to the Koopman operator for an actuated system, and $\bm{C}_{\bm{\theta}} \in \mathbb{R}^{n \times N}$ is a linear decoder. The initial condition of the predictor is obtained by passing the initial condition $\bm{x}_0$ through a nonlinear encoder $\bm{\psi}_{\bm{\theta}}\colon \mathbb{R}^n \mapsto \mathbb{R}^N$, where typically $N \gg n$. All model components can be learned from data by adjusting the model parameters $\bm{\theta}$. When using models of the form given by Equations (\ref{eq:Koopman_model_encoder}) - (\ref{eq:Koopman_model_decoder}) as part of MPC, the nonlinear encoding of the initial state into the higher dimensional Koopman space can be performed before solving the OCP and, therefore, outside of the scope of the OCP optimizer. Given an initial Koopman state $\bm{z}_0 = \bm{\psi}_{\bm{\theta}}(\bm{x}_0)$, the OCP optimizer sees only the linear components of the Koopman model, i.e., $\bm{A}_{\bm{\theta}}$ and $\bm{B}_{\bm{\theta}}$ for the dynamics, and $\bm{C}_{\bm{\theta}}$ to evaluate the objective function or constraints in case they depend on state variables. Thus, all equality constraints imposed by the Koopman model are linear. Therefore, the resulting OCPs are convex if the objective function and constraints on control and system state variables are convex.
    
\subsection{Deep reinforcement learning with continuous action spaces}\label{sec:method_DeepRL}
    Deep RL utilizes nonlinear function approximators, such as deep artificial neural networks, to learn optimal control policies through sequential feedback from trial and error actuation (\cite{sutton2018reinforcement}). The optimal control problem is typically represented by a Markov Decision Process (MDP), wherein the next state $\bm{x}_{t+1}$ and reward $r_{t+1}$ depend only on the current state $\bm{x}_{t}$ and action $\bm{u}_{t}$. The MDP of a specific RL problem is referred to as the environment of that problem. An episode is defined as a single run of an agent in an environment, from the initial state to a terminal state. In engineering applications with continuous action spaces, RL methods that involve agents learning parameterized policies $\bm{\pi}_{\bm{\theta}}(\bm{u}_{t}|\bm{x}_{t})\colon \mathbb{R}^n \mapsto \mathbb{R}^m$, directly mapping states to actions, are the most successful (\cite{sutton2018reinforcement}). These methods can be classified as either policy-gradient or actor-critic methods. In the last decade, actor-critic methods have been the focus of intense research efforts (\cite{silver2014deterministic, fujimoto2018addressing, schulman2015trust, schulman2017proximal, haarnoja2018soft}) and today's state-of-the-art algorithms typically vastly outperform pure policy-gradient methods.

\subsection{Post-optimal sensitivity analysis of convex problems}\label{sec:method_POSA}
    Obtaining post-optimal sensitivities of solution mappings of optimization problems has been a topic of study for several decades (e.g., \cite{fiacco1985sensitivity, ralph1995directional, agrawal2019differentiating}). Given a parametric nonlinear program with parameters $\bm{\theta}$, and a solution point $\bm{a}^{*}$ for a given $\bm{\theta}^{*}$, the solution point is differentiable with respect to the parameters in a neighborhood of $\bm{\theta}^{*}$ under the following assumptions (\cite{pistikopoulos2020multi, fiacco1990sensitivity}): (i) The objective function and constraints of the problem are twice continuously differentiable in $\bm{a}$. (ii) The gradients of objective function and constraints with respect to $\bm{a}$ and the constraints are once continuously differentiable in $\bm{\theta}$. (iii) The second-order sufficient conditions (SOSC) for a local minimum hold at $\bm{a}^{*}$. (iv) The linear independence constraint qualification (LICQ) is satisfied at $\bm{a}^{*}$. (v) Strict complementary slackness (SCS) holds at $\bm{a}^{*}$. Note that the above conditions also imply that the KKT conditions hold (\cite{karush1939minima, boyd2004convex}), and that $\bm{a}^{*}$ is an isolated local minimum (\cite{fiacco1990sensitivity}). For a convex parametric problem this means that the solution map of the optimization problem becomes single-valued, i.e., the solution map is locally an implicit function of the problem parameters.

    A straightforward but costly way to obtain the derivatives of a solution point $\bm{a}^{*}$ that satisfies conditions (i-v) with respect to the parameters $\bm{\theta}$ is to unroll the computational graph of an iterative optimization algorithm and backpropagate through every step of the entire algorithm using algorithmic differentiation (e.g., \cite{domke2012generic}). Alternatively, it is possible to apply the implicit function theorem (\cite{nocedal1999numerical}) to the KKT conditions of the optimization problem to obtain backpropagation gradients for a small fraction of the computational cost required to solve the optimization problem (e.g., \cite{fiacco1990sensitivity}). Using the implicit function theorem, \cite{amos2017optnet} and \cite{Agrawal2019differentiable} connected solvers for quadratic and convex problems, respectively, to the popular deep learning frameworks PyTorch (\cite{paszke2017automatic}) and Tensorflow (\cite{abadi2016tensorflow}). In doing so, they substantially lowered the required effort to use post-optimal sensitivity analysis in a deep learning project.

\subsection{End-to-end learning of Koopman models for MPC}\label{sec:method_e2eKoopman}
    \begin{figure*}[ht]
    	\centering
    	\includegraphics[width=0.65\paperwidth]{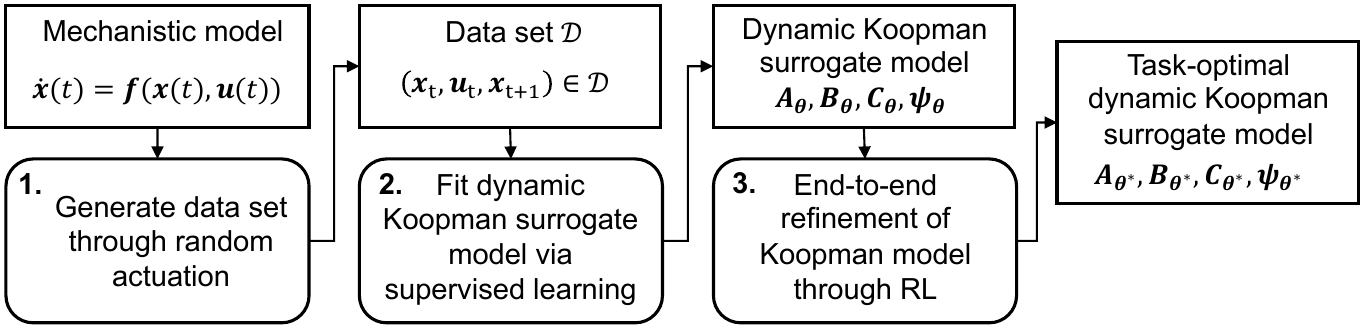}
    	\caption{Workflow from mechanistic model to task-optimal dynamic Koopman surrogate model.}
    	\label{fig:workflow}
    \end{figure*}
    Figure \ref{fig:workflow} summarizes our proposed workflow for training task-optimal Koopman surrogate models. We start with the standard SI procedure for dynamic surrogate models, given a mechanistic model: First, we generate a data set $\mathcal{D}$ from which the system dynamics can be learned using supervised learning, by (randomly) actuating the mechanistic model. Second, we fit a dynamic surrogate model, in our case a Koopman model with learnable parameters $\bm{\theta}$, to the data. As described in Section \ref{sec:method_Koopman4control}, our Koopman models consist of a nonlinear encoder $\bm{\psi_{\theta}}$, the matrices $\bm{A}_{\bm{\theta}}$ and $\bm{B}_{\bm{\theta}}$ that represent the linear dynamics inside the Koopman space, and a decoder matrix $\bm{C}_{\bm{\theta}}$. \cite{lusch2018deep} identify three high-level requirements for autoregressive dynamic Koopman models that correspond to three types of loss functions that need to be combined when performing system identification. The extension to Koopman models for controlled systems and the model structure by \cite{korda2018linear} is straightforward and results in the following requirements and loss functions:
    \begin{enumerate}
        \item Identification of nonlinear lifting functions $\bm{\psi_{\theta}}$ that allow for a linear reconstruction through $\bm{C}_{\bm{\theta}}$. The associated loss function is the autoencoder loss given by
        \begin{equation}\label{eq:Autoencoder_loss}
            || \bm{C_{\theta}} \bm{\psi_{\theta}}(\bm{x}_t) - \bm{x}_t ||.
        \end{equation}
        \item Identification of linear latent space dynamics, using the prediction loss:
        \begin{equation}\label{eq:Predictor_loss}
            || \bm{A}_{\bm{\theta}} \bm{\psi_{\theta}}(\bm{x}_t) + \bm{B}_{\bm{\theta}} \bm{u}_t - \bm{\psi_{\theta}}(\bm{x}_{t+1}) ||
        \end{equation}
        \item State prediction in the original system state, combining all elements of the model. The associated loss function is given by
        \begin{equation}\label{eq:Combined_loss}
            || \bm{C_{\theta}} (\bm{A}_{\bm{\theta}} \bm{\psi_{\theta}}(\bm{x}_t) + \bm{B}_{\bm{\theta}} \bm{u}_t) - \bm{x}_{t+1} ||.
        \end{equation}
    \end{enumerate}
    The norm $||\cdot||$ is the mean-squared error, averaging over the number of samples. Depending on the system, it can be beneficial to use multi-step versions of the prediction loss (Equation \ref{eq:Predictor_loss}) and the combined loss (Equation \ref{eq:Combined_loss}) to improve model accuracy when predicting over multiple time steps in closed-loop fashion. We follow standard SI practices, such as using a separate validation data set for early stopping and normalizing the inputs and outputs of the model.
    
    The main contribution of this work is our method for the third and final step depicted in Figure \ref{fig:workflow}, the end-to-end refinement of Koopman models for optimal performance as part of MPC in a specific control task.
    \begin{figure*}[ht]
    	\centering
    	\includegraphics[width=0.7\paperwidth]{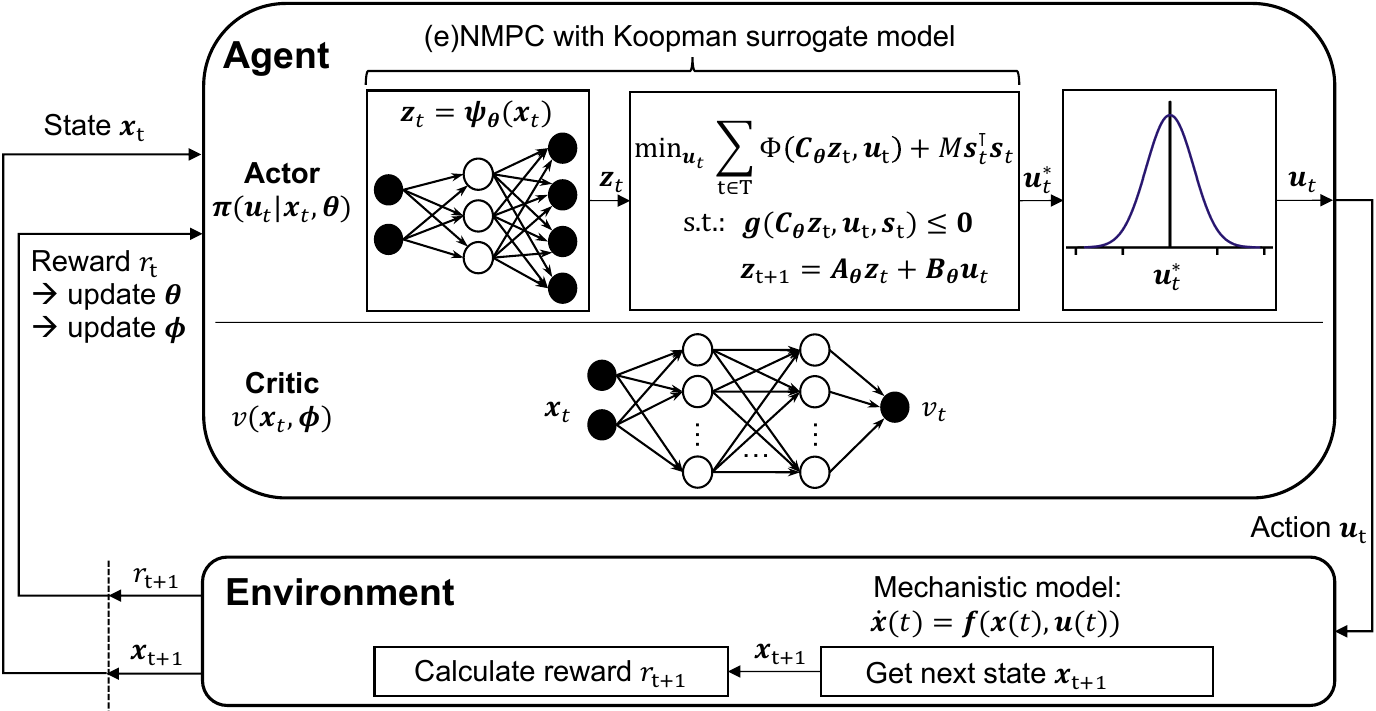}
    	\caption{Method for end-to-end refinement of dynamic Koopman surrogate model. The RL agent consists of a stochastic actor and a critic. The actor is an MPC policy utilizing a dynamic Koopman surrogate model. The critic is a feedforward neural network. The environment consists of the mechanistic model of the system that is to be controlled, and a reward function that depends upon the controllers task.}
    	\label{fig:method}
    \end{figure*}
    Figure \ref{fig:method} summarizes our approach towards end-to-end refinement of Koopman models for optimal performance in (e)NMPC applications through RL. We use the actor-critic PPO algorithm (\cite{schulman2017proximal}) to learn the optimal model parameters $\bm{\theta}$ for a given control task. Note that the refinement applies to all elements of the Koopman model, i.e., the predictor consisting of $\bm{A}_{\bm{\theta}}$ and $\bm{B}_{\bm{\theta}}$, and the autoencoder given by $\bm{\psi_{\theta}}$ and $\bm{C}_{\bm{\theta}}$. Using the \textit{cvxpylayers} package (\cite{Agrawal2019differentiable}), we construct automatically differentiable stochastic Koopman-MPC policies $\bm{\pi}_{\bm{\theta}}(\bm{u}_{t}|\bm{x}_{t})\colon \mathbb{R}^n \mapsto \mathbb{R}^m$ that serve as actors. These policies take a state $\bm{x}_t$ as input and compute the Koopman state $\bm{z}_t$ through a multilayer perceptron encoder $\bm{\psi_{\theta}}$. Then, by evaluating objective function and constraints through the decoder, any (e)MPC problem with a convex objective function and convex constraints (in state space $\bm{x}$) can be formulated and solved using \textit{cvxpylayers} to obtain the optimal control solution $\bm{u}^*_t$. As is common in applying MPC and suitable for an RL approach, our MPC policies are implemented using a receding horizon approach and only return the control solution for the first control step in the MPC horizon. Given the current time step $t$ and an MPC horizon of $t_{f}+1$ steps with the corresponding sets $\mathrm{T}_{+1} = \{ t,\dots,t+t_f \}$ and $\mathrm{T} = \{ t,\dots,t+t_{f}-1 \}$, the OCPs that are solved inside the Koopman-MPC policies take the form
    \begin{subequations}\label{eq:Koopman_OCP}
    	\begin{align}
    		\underset{(\bm{u}_t)_{t \in \mathrm{T}}}{\min} &\sum_{t \in \mathrm{T}_{+1}} \Phi(\bm{C_{\theta}}\bm{z}_t,\bm{u}_t) + M\bm{s}^{\intercal}_{t}\bm{s}_t, \label{eq:OCP_obj_func}\\
    		\text{s.t. }\bm{z}_{t+1} &= \bm{A_{\theta}} \bm{z}_t + \bm{B_{\theta}} \bm{u}_t \quad\forall t \in \mathrm{T}, \label{eq:OCP_state_evolution}\\
    		\underline{\bm{x}} - \bm{s}_t &\leq \bm{C_{\theta}}\bm{z}_t \leq \bar{\bm{x}} + \bm{s}_t \quad\forall t \in \mathrm{T}_{+1}, \label{eq:OCP_state_bounds}\\
    		\underline{\bm{u}} &\leq \bm{u}_k \leq \bar{\bm{u}} \quad\forall t \in \mathrm{T}, \label{eq:OCP_control_bounds}\\
    		\bm{g}(\bm{C_{\theta}}\bm{z}_t,\bm{u}_t,\bm{s}_t) &\leq 0 \quad\forall t \in \mathrm{T}_{+1}. \label{eq:OCP_ineq_constraints}
    	\end{align}
    \end{subequations}
    Both state variables $\bm{x}_t$ and control variables $\bm{u}_t$ can have upper and lower bounds (see Equations (\ref{eq:OCP_state_bounds}) and (\ref{eq:OCP_control_bounds})). To guarantee the feasibility of the OCPs, we add slack variables $\bm{s}_t$ to the state bounds and penalize their use quadratically in the objective function (Equation (\ref{eq:OCP_obj_func})) with a penalty factor $M$. The stage cost $\Phi$ of the objective function can be any convex function and depends on the aim of the policy. The linear transition function in Koopman space is given by Equation (\ref{eq:OCP_state_evolution}). Depending on the application, the OCPs can also include further convex inequality constraints (see Equation \ref{eq:OCP_ineq_constraints}).
    
    As PPO is an on-policy RL method, the actor must be stochastic during training to ensure exploration. Therefore, during training, we sample the action that the actor will take from a normal distribution $\bm{u}_t \sim \bm{\mathcal{N}}(\bm{u}^*_t,\bm{\sigma}^{2})$ with the mean $\bm{u}^*_t$ being the deterministic solution of the Koopman-MPC (see Figure \ref{fig:method}). Depending on the application, it can be beneficial to use a constant standard deviation $\bm{\sigma}$, or start with a large standard deviation and gradually reduce the amount of exploration, e.g., by reducing $\bm{\sigma}$ linearly up to a specific value, as the policy becomes more competent (e.g., \cite{chen2019gnu}).
    
    We use a critic in the form of a multilayer perceptron (MLP) with learnable parameters $\bm{\phi}$ and generalized advantage estimation (\cite{schulman2015high}) to calculate the advantage estimates during training. These are used to compute the actor loss via the clipped PPO loss function and to update the actor parameters $\bm{\theta}$ accordingly (\cite{schulman2017proximal}). We also use global gradient clipping to stabilize the training, as described by \cite{engstrom2020implementation}. We use the Adam optimizer (\cite{kingma2014adam}) to train both the actor and the critic.
    
    We use the running average cumulative reward of a fixed number of preceding episodes as a performance metric during training. We choose the number of episodes to ensure a relatively small variance while limiting the influence of outdated policy parameters due to training updates (cf. Sections \ref{sec:results_CS1} and \ref{sec:results_CS2}). At the end of a training run, we choose the agent that obtained the highest average cumulative reward as the final agent. We then evaluate the performance of that agent by testing it without any exploration, i.e., $\bm{u}_t = \bm{u}^*_t$.

    Due to the convex nature of the OCPs, our Koopman-MPC policies are not only automatically differentiable, but also computationally cheap compared to many alternatives that utilize general nonlinear dynamic models. We expect that Koopman-MPC policies are real-time feasible for a potentially vast range of systems.
    
\section{Numerical experiments}\label{sec:results}

\subsection{Case study description}\label{sec:case_study_description}
    We base our numerical examples on a dimensionless benchmark continuous stirred-tank reactor (CSTR) model (\cite{flores2006simultaneous, du2015time}). The dynamics of the CSTR are given by two nonlinear ordinary differential equations:
    \begin{subequations} 
    	\begin{align}
            \dot{c}(t) = &\:(1-c(t)) \dfrac{\rho(t)}{V} - c(t)ke^{-\frac{N}{T(t)}}, \label{eq:CSTR_model_cdot} \\
            \begin{split}\label{eq:CSTR_model_Tdot}
                \dot{T}(t) = &\:(T_f - T(t))\dfrac{\rho(t)}{V} + c(t)ke^{-\frac{N}{T(t)}}\\
                &\:- F(t) \alpha_c (T(t)-T_c)
            \end{split}
    	\end{align}
    \end{subequations}
    The system states $\bm{x}$ are the product concentration $c$ and the temperature $T$. The control inputs $\bm{u}$ are the production rate $\rho$ and the coolant flow rate $F$. All parameters are given in Table \ref{tab:CSTR_parameters}.
    \begin{table}[ht]
    \centering
    \setlength{\extrarowheight}{0.05cm}
        \caption{CSTR model parameters (\cite{flores2006simultaneous, du2015time})}
        \label{tab:CSTR_parameters}
        \begin{tabular}{lll}
            \toprule
                                      &    symbol  & value              \\ \hline
            volume                    &      $V$   & $20$               \\
            reaction constant         &      $k$   & $300\frac{1}{h}$   \\
            activation energy         &      $N$   & $5$                \\
            feed temperature          &      $T_f$ & $0.3947$           \\
            heat transfer coefficient & $\alpha_c$ & $1.95\cdot10^{-4}$   \\
            coolant temperature       &      $T_c$ & $0.3816$             \\
            \bottomrule
        \end{tabular}
    \end{table}
    
    Based on the CSTR model, we construct both NMPC and eNMPC applications: In NMPC (Section \ref{sec:results_CS1}), the controller shall stabilize the product concentration $c$ and temperature $T$ at given target values, given random perturbations to the production rate $\rho$. The goal in our eNMPC case (Section \ref{sec:results_CS2}) is to optimize process economics while satisfying process constraints, given electricity price predictions. Table \ref{tab:CSTR_bounds_and_target} presents the bounds of the system states and control inputs as well as the target values of the system states in NMPC.
    \begin{table*}[ht]
    \centering
    \setlength{\extrarowheight}{0.05cm}
        \caption{Lower (lb) and upper (ub) bounds of system states and control inputs, target values of system states in NMPC, and steady-state (ss) values used to evaluate the economic benefit of flexible production in eNMPC.}
        \label{tab:CSTR_bounds_and_target}
        \begin{tabular}{l|rrr|rrr}
            \toprule
            \multicolumn{1}{l|}{} & \multicolumn{3}{c|}{NMPC (Section \ref{sec:results_CS1})} & \multicolumn{3}{c}{eNMPC (Section \ref{sec:results_CS2})} \\
            variable              & lb                  & ub                    & target      & lb                & ub                  & ss        \\ \hline
            $c$                   & -                   & -                     & 0.1367      & 0.1231            & 0.1504              & 0.1367    \\
            $T$                   & -                   & -                     & 0.7293      & 0.6               & 0.8                 & 0.7293    \\
            $\rho$                & $0.8\frac{1}{h}$    & $1.2\frac{1}{h}$      & -           & $0.8\frac{1}{h}$  & $1.2\frac{1}{h}$    & $1.0\frac{1}{h}$ \\
            $F$                   & $0.0\frac{1}{h}$    & $700.0\frac{1}{h}$    & -           & $0.0\frac{1}{h}$  & $700.0\frac{1}{h}$  & $390.0\frac{1}{h}$ \\
            \bottomrule
        \end{tabular}
    \end{table*}

    We use discrete time steps of length $\Delta t_{\text{discr}} = \SI{15}{\text{min}}$ and control steps of length $\Delta t_{\text{ctrl}} = \SI{60}{\text{min}}$. Given a current state $\bm{x}_t$ and control action $\bm{u}_t$, we use Equations (\ref{eq:CSTR_model_cdot}) - (\ref{eq:CSTR_model_Tdot}) and the RK45 solver (\cite{dormand1980family}) in SciPy (\cite{virtanen2020scipy}) to calculate the next state $\bm{x}_{t+1}$ in the environment.
    
    We compare the performances of three different controller types:
    \begin{enumerate}
        \item \textit{Koopman-SI}: MPC controller using a Koopman model to approximate the CSTR dynamics. The model is trained using SI on a data set generated using the mechanistic model (Equations (\ref{eq:CSTR_model_cdot}) - (\ref{eq:CSTR_model_Tdot})) and random control inputs. We provide a more detailed explanation of the data sampling and SI procedure in Subsection \ref{sec:results_data_sampling_and_SI}.
        \item \textit{Koopman-RL}: MPC controller using a Koopman model to approximate the CSTR dynamics. We use the model of the \textit{Koopman-SI} controller as an initial guess and tune it for task-optimal MPC performance using the approach described in Section \ref{sec:method_e2eKoopman}.
        \item \textit{MLP}: A model-free neural network policy $\bm{\pi}_{\bm{\theta}}(\bm{u}_t|\bm{x}_t)$ in the form of a multilayer perceptron (MLP), trained by RL. Similarly to the \textit{Koopman-RL}, each action is sampled from a normal distribution $\bm{u}_t \sim \bm{\mathcal{N}}(\bm{u}^*_t,\bm{\sigma}^{2})$, where the neural network produces $\bm{u}^*_t$ given the current state of the environment.
    \end{enumerate}
    
    Matching our proposed workflow (see Figure \ref{fig:workflow}), the remainder of this section is structured as follows: Subsection \ref{sec:results_data_sampling_and_SI} elaborates on the data sampling procedure using the mechanistic model and discusses the subsequent SI process using the Koopman model architecture. Subsections \ref{sec:results_CS1} and \ref{sec:results_CS2} describe the end-to-end refinement of the Koopman models and report the performances of all three controller types for the NMPC and eNMPC applications, respectively. Finally, Subsection \ref{sec:results_CS2_2} presents our findings on a modified eNMPC, which we use to study the adaptability of end-to-end refined Koopman MPCs to modifications of constraints after training has been completed.

\subsection{Data sampling and system identification}\label{sec:results_data_sampling_and_SI}
    Given randomly sampled control inputs $\bm{u}$ within their respective bounds, the product concentration $c$ typically quickly moves away from its feasible range in eNMPC and also away from its target value in NMPC (see Table \ref{tab:CSTR_bounds_and_target}). Therefore, such a sampling strategy would not produce a data set particularly suitable for learning a good dynamic surrogate model, as most data would be in irrelevant parts of the state space. Instead of actuating the mechanistic model by randomly sampling both control inputs $\rho$ and $F$, we, therefore, take a different approach: We generate multiple trajectories using the mechanistic model, each trajectory starting from the target state of the NMPC, which is also at the center of the feasible range of the eNMPC. For each trajectory, we randomly generate a series of control inputs for either $\rho$ or $F$. Given this series of inputs, we then solve an OCP using the mechanistic model to determine a corresponding series of values for the respective other control variable, which leads to $c$ not diverging much from its feasible range in eNMPC. As in NMPC and eNMPC, we use control steps of length $\Delta t_{\text{ctrl}} = \SI{60}{\text{min}}$, i.e., $\rho$ and $F$ change every four time steps.
    
    Given an OCP horizon of $t_f$ time steps with the corresponding set $\mathrm{T} = \{ 0,\dots,t_{f}-1 \}$, and a randomly generated trajectory for the production rate $\rho_{\text{target},t} \forall t \in \mathrm{T}$, the OCP minimizes
    \begin{equation}\label{eq:OCP_datagen_roh_setpoint}
        \sum_{t \in \mathrm{T}} (c_{\text{target}} - c_t)^2 + w_{\rho} (\rho_{\text{target},t} - \rho_t)^2,
    \end{equation}
    subject to the mechanistic model (Equations (\ref{eq:CSTR_model_cdot}) - (\ref{eq:CSTR_model_Tdot})), and subject to the given bounds on control variables (see Table \ref{tab:CSTR_bounds_and_target}), with $c_{\text{target}}$ being the target value for $c$ in NMPC (see Table \ref{tab:CSTR_bounds_and_target}) and $w_{\rho} = 10$. Equivalently, given a sampled trajectory for the coolant flow rate $F_{\text{target},t} \forall t \in \mathrm{T}$, the OCP minimizes
    \begin{equation}\label{eq:OCP_datagen_F_setpoint}
       \sum_{t \in \mathrm{T}} (c_{\text{target}} - c_t)^2 + w_{F} \cdot (F_{\text{target},t} - F_t)^2,
    \end{equation}
    with $w_F = 0.1$. We choose $w_{\rho}$ and $w_F$ so that the resulting data set has good coverage of the feasible part of the state space (see Table \ref{tab:CSTR_bounds_and_target}, eNMPC). We use the GEKKO Optimization Suite (\cite{beal2018gekko}) with default settings to formulate and solve the OCPs. Given the resulting trajectories for the control inputs $\bm{u}$, we use the RK45 solver (\cite{dormand1980family}) in SciPy (\cite{virtanen2020scipy}) to compute the system response and generate the data set for SI. We generate 84 trajectories for training and validation purposes. Each of those trajectories has a length of 5 days, i.e., 480 time steps. We use 63 of those trajectories for training and the remaining 21 for validation.

    We perform SI by minimizing the sum of three loss functions (see Equations \ref{eq:Autoencoder_loss} - \ref{eq:Combined_loss}), corresponding to the three requirements to a dynamic Koopman model (see Section \ref{sec:method_e2eKoopman}). To obtain Koopman models that perform well over multiple time steps of closed-loop prediction, we take a stochastic curriculum learning approach (\cite{Bengio2015_curriculum, zaremba2014learning}), i.e., we start the training procedure by minimizing the mean-squared error (MSE) of predictions over a single time step, and, progress onto minimizing the average MSE over multiple time steps of continuous closed-loop prediction in later epochs. Such a curriculum learning procedure is designed to result in models that make accurate long-term predictions but to avoid exploding/vanishing gradients resulting from concatenating predictions of an untrained model in the first epochs. Specifically, in the first epoch, we use the one-step MSE as the loss function. After 250 epochs, we use the average MSE over 240 time steps, i.e., 2.5 days, of closed-loop prediction in every epoch. Between the first and the 250th epoch, we decrease the probability of taking the one-step loss linearly from \SI{100}{\%} to \SI{0}{\%} in favor of the 240-step loss. We use the Adam optimizer (\cite{kingma2014adam}) with a learning rate of $0.5\cdot10^{-4}$, a mini-batch size of 64 samples, and a maximum number of 5,000 epochs. We stop training early after at least 350 epochs if the validation loss has not reached a new minimum for 100 consecutive epochs.
    
    A pragmatic and common practice when training ML models is to oversize the models and then use regularization techniques and early stopping to prevent overfitting (\cite{Goodfellow-et-al-2016}). However, we aim to obtain models that strike a good balance between accuracy and computational tractability. Therefore, we avoid unnecessary oversizing of our Koopman surrogate models by starting with a small model and repeatedly training larger models until the performance gains become negligible. All results shown in the following are obtained using a Koopman model with a latent space dimensionality of eight, i.e., $\bm{A}_{\bm{\theta}} \in \mathbb{R}^{8 \times 8}, \bm{B}_{\bm{\theta}} \in \mathbb{R}^{8 \times 2}, \bm{C}_{\bm{\theta}} \in \mathbb{R}^{2 \times 8}$, and an MLP encoder $\bm{\psi}\colon \mathbb{R}^2 \mapsto \mathbb{R}^8$ with two hidden layers (four and six neurons, respectively), and hyperbolic tangent activation functions.

\subsection{NMPC}\label{sec:results_CS1}
    Here, the controller task is to stabilize the product concentration $c$ and temperature $T$ at constant target values (see Table \ref{tab:CSTR_bounds_and_target}). Every eight hours, the production rate $\rho$ is set to a random value within its bounds for the next eight hours. The controller can respond by varying the coolant flow rate $F$ within its bounds to stabilize $c$ and $T$. Thus, the production rate becomes an external input $\rho_{\text{ext}}$ and the coolant flow rate $F$ is the control input.
    
    During RL, we define the reward $r_t$ for each control action as proportional to the negative squared relative deviation between $c$, $T$, and their respective steady-state values:
    \begin{equation}\label{eq:reward_CS1} 
        \begin{split}
                r_t = &\:- ((c_t - c_{\text{target}}) / (c_{\text{ub}} - c_{\text{lb}}))^2\\
                &\:- ((T_t - T_{\text{target}}) / (T_{\text{ub}} - T_{\text{lb}}))^2
        \end{split}
    \end{equation}
    Relative deviations are achieved by normalizing each variable using the size of its feasible range (see Table \ref{tab:CSTR_bounds_and_target}). Aptly, using the feasible range of each variable, we also normalize the inputs and outputs of the Koopman models to values between zero and one. At each control step, the MPC controller solves an OCP in Koopman space, minimizing the squared distance from the target state $\bm{x}_{\text{target}}$:
    \begin{subequations}\label{eq:Koopman_OCP_CS1}
    	\begin{align}
    		\underset{(F_t)_{t \in \mathrm{T}}}{\min} &\sum_{t \in \mathrm{T}_{+1}} || C_{\bm{\theta}}\bm{z}_t - \bm{x}_{\text{ss}} ||,\label{eq:Koopman_OCP_CS1_objfunc}\\
    		\text{s.t. }\bm{z}_0 &= \bm{\psi}_{\bm{\theta}} (\bm{x}_0),\\
    		\bm{z}_{t+1} &= A_{\bm{\theta}}\bm{z}_t + B_{\bm{\theta}}\colvec{2}{\rho_{\text{ext}}}{F_t} \; &\forall t \in \mathrm{T},\\
    		\underline{F}_t &\leq F_t \leq \bar{F}_t \; &\forall  t \in \mathrm{T}.
    	\end{align}
    \end{subequations}
    We use an MPC horizon of $t_f = 12$ quarter hours, i.e., three hours, and episodes with a duration of 288 time steps, i.e., three days. As the production rate $\rho_{\text{ext}}$ is adjusted every eight hours, each episode includes eight step changes.
    
    We repeat the RL training of the \textit{Koopman-RL} and \textit{MLP} controllers over ten random seeds. The values of training hyperparameters are given in the supplementary material. The MLP controllers have two fully connected hidden layers, each with 256 neurons, followed by an output layer with one neuron for $F$. All layers have hyperbolic tangent activation functions. As a performance indicator during the training process, we use the running average sum of rewards during the previous 30 episodes, i.e., 2,160 control steps. As the policies are updated every 2,048 control steps, taking the running average of 30 episodes has a relatively low variance while not being overly strongly influenced by actions resulting from outdated policy parameters. In the following, we call the undiscounted cumulative reward of an episode the \textit{score} of that episode.
    \begin{figure*}[htb] 
    	\centering
    	\includegraphics[width=0.7\paperwidth]{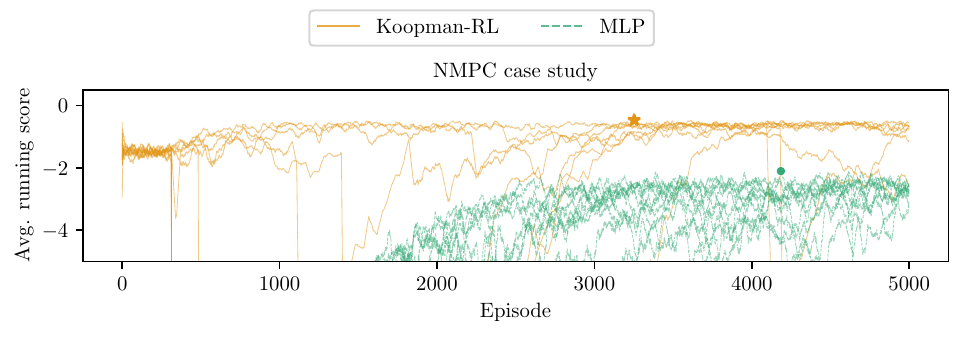}
    	\caption{Summary of training runs in NMPC. Each training configuration is run ten times with different random seeds. We average the running score over the last 30 episodes. The lines represent individual training runs. Additionally, we highlight the highest average running score achieved by a policy type (star for \textit{Koopman-RL}, dot for \textit{MLP}).}
    	\label{fig:training_results_NMPC}
    \end{figure*}
    
    Figure \ref{fig:training_results_NMPC} summarizes the RL training results for both RL-trained controller types. We show the running average of training scores of each training run. Due to using a pretrained Koopman model, the \textit{Koopman-RL} controllers start the RL training at comparatively high scores. Still, after around 5,000 episodes, most \textit{Koopman-RL} controllers have significantly increased their average scores compared to the SI baseline, and the performance reaches a plateau. Some training runs show rather steadily improving scores from the start, reaching the plateau after about 1,000 episodes. Other trainings show an initial drop in performance, followed by a recovery. The MLP controllers are randomly initialized and obtain low scores at the beginning of the training. The performance improves gradually and plateaus after about 3,000 episodes at average scores below those of the SI-pretrained Koopman controllers.
    
\begin{figure*}[htb]
    	\centering
    	\includegraphics[width=0.7\paperwidth]{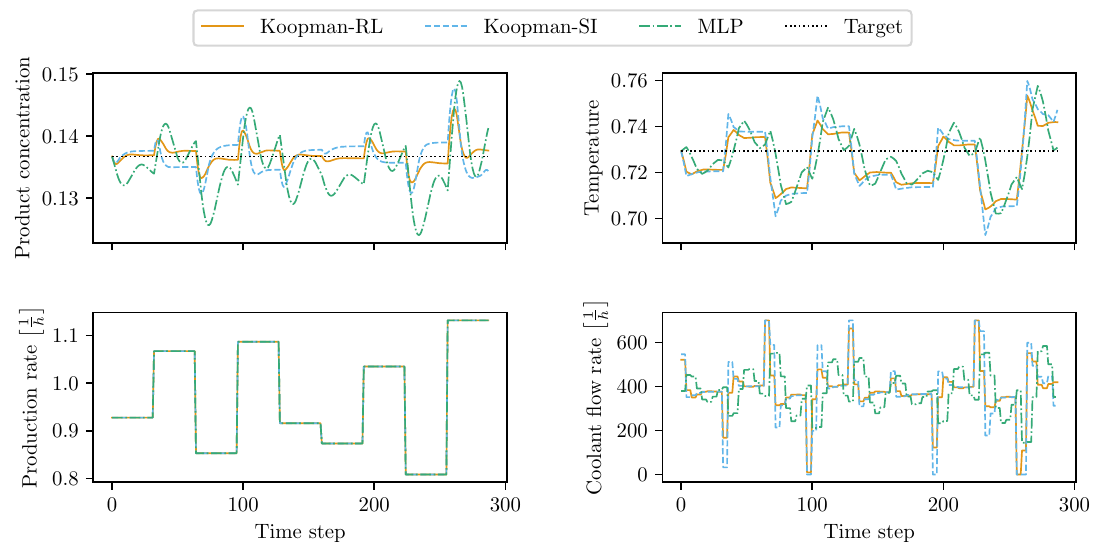}
    	\caption{NMPC: comparison of the controller behavior, given the same randomly generated production rate trajectory. Best viewed in color.}
    	\label{fig:trajectory_comparison_CS1}
    \end{figure*}

    After completing all training runs, we test the policies that achieved the highest average running scores without exploration, i.e., we set $\bm{u}_t = \bm{u}^*_t$, on 100 test episodes. Table \ref{tab:cs1_performance} shows the results of this test. A direct comparison of the three controllers, given the same randomly generated production rate trajectory, is given in Figure \ref{fig:trajectory_comparison_CS1}. Note that as exploration is turned off in these tests, the performance may be higher than during training (cf. Figure \ref{fig:training_results_NMPC}).
    \begin{table}[htb] 
    \centering
    \setlength{\extrarowheight}{0.05cm}
        \caption{NMPC: test results over 100 episodes. Scores represent the negative summed squared relative deviation from the target values, thus higher is better and a score of 0 is the upper performance bound.}
        \label{tab:cs1_performance}
        \begin{tabular}{lllll}
            \toprule
            Score               & avg.     & std.  & min.  & max.   \\ \hline
            \textit{MLP}        & -2.21    & 0.73  & -4.40 & -0.43  \\
            \textit{Koopman-SI} & -1.35    & 0.62  & -3.06 & -0.29  \\
            \textit{Koopman-RL} & -0.50    & 0.19  & -1.27 & -0.12  \\
            \bottomrule
        \end{tabular}
    \end{table}
    
    All three controllers exhibit sensible behavior. However, Table \ref{tab:cs1_performance} shows that the \textit{Koopman-RL} controller manages to keep the state of the CSTR on average closer to the target than the other two controllers. This finding is confirmed by Figure \ref{fig:trajectory_comparison_CS1}, which shows that the \textit{Koopman-RL} controller manages to stabilize the product concentration $c$ quicker and much closer to the target than the other two controllers. The \textit{MLP} controller does not stabilize $c$ at or near the target; instead, it makes $c$ oscillate around its target. Regarding the temperature variable $T$, all three controllers produce an average deviation from $T_{\text{target}}$ that is substantially larger than the average deviation of $c$ from $c_{\text{target}}$. This has two reasons: Firstly, the feasible range of $T$ is larger than that of $c$ (see Table \ref{tab:CSTR_bounds_and_target}), which affects the reward calculation (Eq. \ref{eq:reward_CS1}) as we normalize deviations using the feasible range of each variable. Secondly, the objective function in the Koopman OCPs (Eq. \ref{eq:Koopman_OCP_CS1_objfunc}) is also affected due to the normalization of the inputs of the Koopman models. The overall performance regarding $T$ is very similar for all three controllers. Again, the $MLP$ controller produces oscillating behavior, whereas both Koopman controllers establish steady states at some distance to the target after each step change.

\subsection{eNMPC}\label{sec:results_CS2}
    Here, the controller task is to optimize process economics through demand response, i.e., shifting electricity consumption in time to benefit from intervals with relatively low prices, given electricity price profiles. The state variables $c$ and $T$ are subject to box constraints, with upper and lower bound values given in Table \ref{tab:CSTR_bounds_and_target}. The control inputs $\bm{u}_t = (\rho_t, F_t)^\intercal$ can be varied freely within their bounds. For simplicity, we assume that electric power consumption is proportional to the coolant flow rate $F$. Thus, the controller can improve the process economics by shifting process cooling to low-price intervals. To enable flexible operation, we assume a product storage with filling level $l_t$ and a maximum capacity of six hours of steady-state production. Because the number of variables in an OCP scales with the prediction horizon, RL training with a differentiable MPC controller and a long time horizon can quickly become computationally infeasible. We identify a price prediction horizon that results in a performant eNMPC by repeatedly solving eNMPC control problems using the mechanistic model (Equations (\ref{eq:CSTR_model_cdot}) - (\ref{eq:CSTR_model_Tdot})). Based on the results, we choose a price prediction horizon of nine hours, as longer horizons yield no substantial improvement in control performance.
    
    We use historic day-ahead electricity prices from the Austrian market (\cite{open_power_system_data_2020}) from March 29, 2015 to March 25, 2018 to train the controllers. The final testing is performed on Austrian electricity price data from March 26, 2018 to September 30, 2018. Our goal is to train controllers that perform well when used continuously over an arbitrarily long time horizon. In addition, when testing controllers on short episodes in the eNMPC setting, the economic performance of a controller can be heavily influenced by how the controller utilizes the product storage. For example, a controller that fills up the storage initially to obtain flexibility might incur higher costs in a short episode than a steady-state production but might achieve economic savings in the long run. Therefore, we test the final performance of the controllers in a single continuous episode that covers half a year. However, during training, it makes sense to use shorter episodes to allow for more frequent restarts. This is especially important at the beginning of the training procedure when controllers are likely to steer the system to undesirable states from which little can be learned. Therefore, as in the NMPC case study, we choose an episode length of 288 control steps, i.e., three days, for training. At the start of each training episode, the state variables get reset to their steady-state values, the storage level is initialized randomly between one and two hours of steady-state production volume, and a connected series of electricity prices of length $\mathrm{T}$ is sampled from the data. In the roughly half-year-long test episode, we start with an empty storage to have a consistent starting point for the three different controllers. The product concentration $c$ and the temperature $T$ are initialized at their steady-state values.
    
    For RL, we choose a reward that depends on the electricity cost savings compared to steady-state production (see Table \ref{tab:CSTR_bounds_and_target}) and, on whether any bounds on state variables were violated during the previous control interval. First, we check for constraint violations, and compute the steady-state costs
    \begin{equation}\label{eq:ss_cost}
        c_{\text{ss}} = F_{\text{ss}} \cdot p_t \cdot \Delta t_{\text{ctrl}},
    \end{equation}
    and the costs induced by the current control action
    \begin{equation}\label{eq:actual_cost}
        c_t = F_t \cdot p_t \cdot \Delta t_{\text{ctrl}},
    \end{equation}
    using the electricity price $p_t$. Then, we calculate the reward
    \begin{equation}\label{eq:reward_CS2}
        r_t = 
        \begin{cases}
            -1,                             & \text{if constraint violation}\\
            (c_{\text{ss}} - c_t) \beta,    & \text{otherwise},
        \end{cases}
    \end{equation}
    wherein $\beta$ is a hyperparameter used to balance the influence of cost savings and constraint violations.
    
    At each control step, the eMPC controller solves an OCP in Koopman space, aiming to minimize the operational costs while satisfying all constraints:
    \begin{subequations}\label{eq:Koopman_OCP_CS2}
    \allowdisplaybreaks
	\begin{align}
		\underset{(\rho_t, F_t)_{t \in \mathrm{T}}}{\min} \sum_{t \in \mathrm{T}_{+1}} (F_t p_t \Delta t_{\text{ctrl}} + M\bm{s}^{\intercal}_{t}\bm{s}_t),
	\end{align}
	\vspace{-0.5cm}
	\begin{align}
		\text{s.t. }\bm{z}_0 &= \bm{\psi}_{\bm{\theta}} \colvec{2}{c_0}{T_0},&\label{eq:Koopman_OCP_CS2_encode}\\
		\bm{z}_{t+1} &= \bm{A}_{\bm{\theta}}\bm{z}_t + \bm{B}_{\bm{\theta}}\bm{u}_t \quad\forall t \in \mathrm{T},\\
		l_{t+1} &= l_t + (\rho_t - \rho_{\text{ss}}) \Delta t_{\text{discr}} \quad\forall t \in \mathrm{T},\\
		\bm{x}_{t} &= \bm{C}_{\bm{\theta}}\bm{z}_t \quad\forall t \in \mathrm{T}_{+1},\label{eq:Koopman_OCP_CS2_decode}\\
		\underline{\bm{x}}_t - \bm{s}_{\bm{x},t} &\leq \bm{x}_t \leq \bar{\bm{x}}_t + \bm{s}_{\bm{x},t} \quad\forall  t \in \mathrm{T}_{+1},\label{eq:Koopman_OCP_CS2_StateBounds}\\
		0 - s_{l,t} &\leq l_t \leq 6.0 + s_{l,t} \quad\forall  t \in \mathrm{T}_{+1},\\
		\bm{s}_t &= \colvec{2}{\bm{s}_{\bm{x},t}}{s_{l,t}} \quad\forall  t \in \mathrm{T}_{+1},\\
		\bm{0} &\leq \bm{s}_t \quad\forall  t \in \mathrm{T}_{+1},\\
		\underline{\bm{u}}_t &\leq \bm{u}_t \leq \bar{\bm{u}}_t \quad\forall  t \in \mathrm{T}.
	\end{align}
    \end{subequations}
    As explained in Section \ref{sec:method}, we introduce slack variables $\bm{s}_{\bm{x},t}$ for the state variables and $s_{l,t}$ for the product storage to ensure feasibility of the policy but penalize their use in the objective function.
    
    The values of the training hyperparameters are given in the supplementary material. The MLP controllers have separate input layers for $\bm{x}_{t}$, $l_t$, and the trajectory of future electricity prices. Each of those input layers is followed by a hidden layer, with 100, 56, and 100 neurons, respectively. The outputs of those layers are concatenated and passed through two fully connected layers, each of size 256 neurons. The output layer has two neurons, one each for $\rho$ and $F$. All layers have hyperbolic tangent activation functions.
   
\begin{figure*}[ht]
    	\centering
    	\includegraphics[width=0.7\paperwidth]{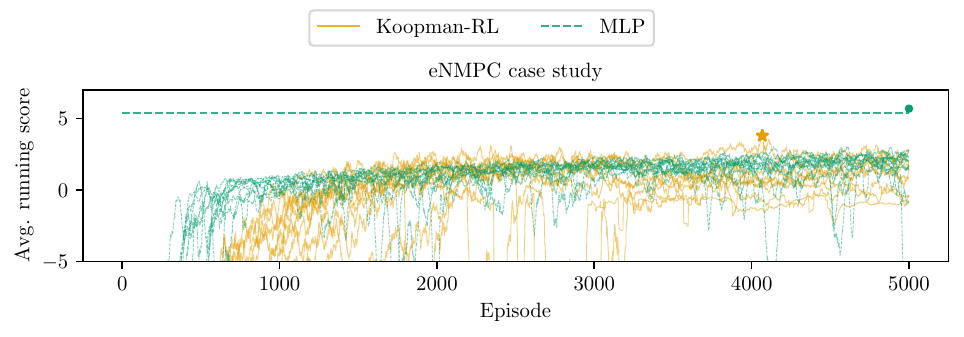}
    	\caption{Summary of RL training runs in eNMPC. Each training configuration is run ten times with different random seeds. We average the running score over the previous 30 episodes. The lines represent individual training runs. The constant dashed green line represents the median maximum performance when training the MLP policies for 20,000 episodes. Additionally, we highlight the highest average running score achieved by a policy type (star for \textit{Koopman-RL}, dot for \textit{MLP}). Best viewed in color.}
    	\label{fig:training_results_eNMPC}
    \end{figure*}

    As in NMPC, we repeat the RL training of the \textit{Koopman-RL} and \textit{MLP} controllers over ten random seeds. Again, we use the average running score during the previous 30 episodes, i.e., 2,160 control steps, as a performance indicator during the training process. Figure \ref{fig:training_results_eNMPC} summarizes the RL training results for both controller types. We show the running average of training scores of each training run. Unlike in NMPC, the SI pretraining of the Koopman models does not lead to high initial scores, as the resulting Koopman MPCs frequently produce minor constraint violations resulting in low rewards (see Equation \ref{eq:reward_CS2}). The performance of the \textit{Koopman-RL} controllers improves and saturates after about 2,500 episodes. After 5,000 episodes, the maximum running scores of the \textit{Koopman-RL} controllers have reached values between $1.4$ and $3.8$. The running scores of the \textit{MLP} controllers rise to around zero after only about 750 episodes. Afterward, the running scores continue to rise slowly, saturating only after about 20,000 episodes at maximum running scores between $4.6$ and $5.7$. We stopped the RL refinement of the \textit{Koopman-RL} controllers at 5,000 episodes because of the high computational cost of the training procedure compared to the \textit{MLP} controllers, which is due to the necessity to solve and backpropagate through OCPs repeatedly. Because our case study is relatively small, analyzing training runtimes would give little insight into the expected runtime for a practical control problem. Thus, we defer this analysis for future work on more realistic systems.
    
    We observe that convergence to high average scores is unstable, especially for the Koopman-RL policies. This is a common finding in policy optimization algorithms (\cite{schulman2015trust, schulman2017proximal}), stemming from the fact that small perturbations to the parameters $\bm{\theta}$ of a policy can have outsized effects on the resulting actions, and thus also the rewards. We hypothesize that this effect is aggravated when learning the parameters of an MPC policy, as small changes in trajectory predictions can lead to very different active sets of constraints, resulting in vastly different control outputs.
    
\begin{figure*}[ht]
    	\centering
    	\includegraphics[width=0.7\paperwidth]{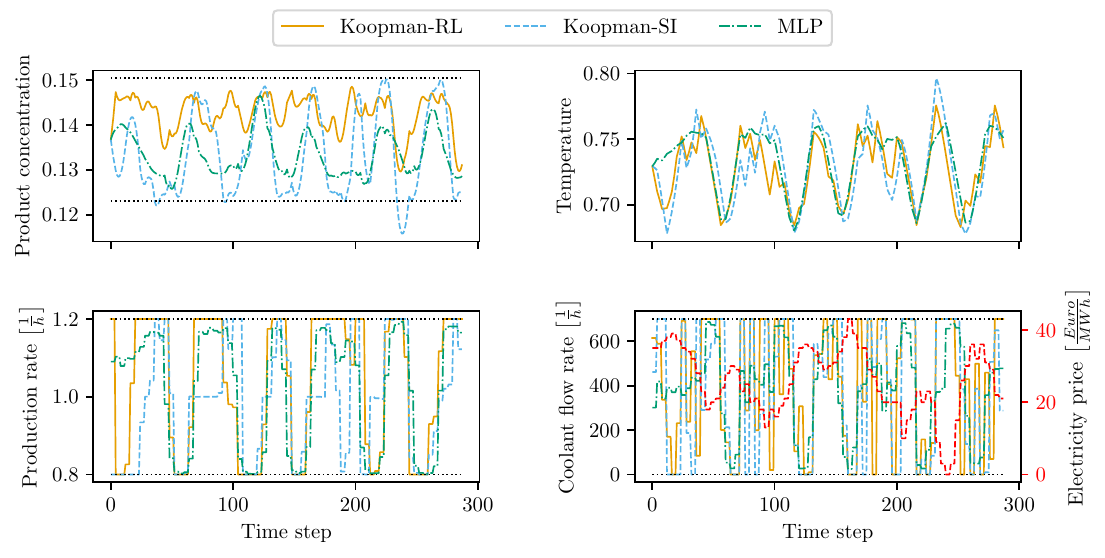}
    	\caption{eNMPC: comparison of the controller behavior, given a randomly sampled continuous electricity price trajectory from the test set. The dotted lines represent the bounds (Table \ref{tab:CSTR_bounds_and_target}). Best viewed in color.}
    	\label{fig:trajectory_comparison_CS2}
    \end{figure*}

    Similar to our analysis in Section \ref{sec:results_CS1}, after completing the controller training runs, we test the policies which achieved the highest ever running scores during training. We perform this testing without exploration, i.e., we set $\bm{u}_t = \bm{u}^*_t$, and we report the performance metrics in Table \ref{tab:cs2_performance}. Additionally, we show the control behavior of the three controllers in a three-day test episode using a randomly sampled continuous electricity price profile from the test set in Figure \ref{fig:trajectory_comparison_CS2} for illustration purposes.
    \begin{table*}[ht]
        \centering
        \setlength{\extrarowheight}{0.05cm}
        \caption{eNMPC: test results over a single roughly half-year long test episode with electricity price data from March 26, 2018 to September 30, 2018. Costs are stated relative to steady-state production. The constraint violation column reports the percentage of control steps that result in constraint violations.}
        \label{tab:cs2_performance}
        \begin{tabular}{l|rrr}
            \toprule
                                & Cost & Constr. viols. [\%] & Avg. storage level [h] \\ \hline
            \textit{MLP}        & 0.91 &  0.24               & 0.80               \\
            \textit{Koopman-SI} & 0.90 &  8.84               & 0.28               \\
            \textit{Koopman-RL} & 0.94 &  0.22               & 5.32               \\
            \bottomrule
        \end{tabular}
    \end{table*}
    
    Table \ref{tab:cs2_performance} shows that even though the \textit{Koopman-SI} controller achieves the highest cost savings, it does so at the expense of causing many constraint violations. The \textit{Koopman-RL} controller and the \textit{MLP} controller cause few constraint violations. In case of the \textit{Koopman-RL} controller, all violations are related to the bounds of the concentration variable $c$. The \textit{MLP} controller achieves higher cost savings than the \textit{Koopman-RL} controller. Interestingly, all constraint violations by the \textit{MLP} controller are violations of the non-negativity constraint of the product storage. Apparently, the controllers utilize the product storage quite differently. While the \textit{Koopman-SI} controller mostly keeps the storage near-empty, the \textit{Koopman-RL} controller exhibits the opposite behavior, generally operating close to the maximum storage capacity of six hours, which could partially explain the difference in economic performance compared to the \textit{MLP} controller. Most importantly, end-to-end training clearly improved the overall performance of the Koopman MPC in this application.

\subsubsection{Analysis of Koopman embedding before and after end-to-end training}\label{sec:results_analysis_koopman_embedding}
    A central challenge in applied Koopman theory is identifying a suitable set of functions for mapping between the original state space and the lifted space where the dynamics can be approximated linearly with high accuracy. Here, we analyze how the proposed end-to-end refinement of a Koopman model for task-optimal performance in eNMPC affects the autoencoder part of the model, which defines the mapping between the original and lifted domains and which is pre-trained by SI and then fine-tuned by RL. To this end, we generate a data set of 100,000 steps in the demand response environment using the full mechanistic model by applying the \textit{Koopman-SI} controller. This data set contains 8,302 steps with constraint violations in either $c$ or $T$, and 91,698 steps without constraint violations. We use this data set to calculate the autoencoder MSE for both the \textit{Koopman-SI} and the \textit{Koopman-RL} model as a measure for the reconstruction loss. Furthermore, we pass the 100,000 points through the autoencoders of both models. Then, we calculate the rates at which points with violations of state bounds get correctly mapped to the exterior of the feasible region defined by the state bounds (sensitivity). To determine the specificity, we proceed analogously for the points without constraint violations.

    \begin{table*}[h]
    \centering
    \setlength{\extrarowheight}{0.05cm}
    \caption{Analysis of autoencoder performance and constraint violation detection rates for the \textit{Koopman-SI} and \textit{Koopman-RL} models. The table displays the autoencoder mean-squared error (MSE) as a measure for the reconstruction loss, sensitivity (true positive rate), and specificity (true negative rate) for different combinations of encoder and decoder from \textit{Koopman-SI} (SI) and \textit{Koopman-RL} (RL) models. Herein, \textit{positive} refers to a point with a constraint violation in $c$ or $T$, and \textit{negative} to a point without a constraint violation.}
    \label{tab:koopman_embedding_analysis}
        \begin{tabular}{cc|ccc}
        \toprule
        \makecell{Encoder\\model} & \makecell{Decoder\\model} & \makecell{Autoencoder\\MSE} & \makecell{True pos.\\(sensitivity)} & \makecell{True neg.\\(specificity)} \\ \hline
        SI            & SI            & $3.2\cdot10^{-5}$   & 0.844     & 0.995   \\
        RL            & RL            & $3.5\cdot10^{-4}$   & 0.980     & 0.947   \\
        \hline
        SI            & RL            & $3.5\cdot10^{-5}$   & 0.844     & 0.995   \\
        RL            & SI            & $4.1\cdot10^{-4}$   & 0.971     & 0.947   \\
        \bottomrule
        \end{tabular}
    \end{table*}

    As can be seen from the upper part of Table \ref{tab:koopman_embedding_analysis}, the end-to-end refinement leads to a tenfold increase in the reconstruction loss; still, the loss remains at a relatively low level. Importantly, the autoencoder of the \textit{Koopman-RL} model shows a much higher sensitivity than the autoencoder of the \textit{Koopman-SI} model, i.e., the rate at which a point that violates a constraint on $c$ or $T$ still does so after being passed through the autoencoder rises from $84.4\;\%$ to $98.0\;\%$. When using the Koopman models as part of an eNMPC, this increase in sensitivity and the related decrease in specificity leads to more conservative controller behavior. The higher sensitivity and specificity are expected as the Koopman model was refined using a reward function that heavily punishes constraint violations (see Eq. \ref{eq:reward_CS2}). 

    To identify which part of the autoencoder model was modified to what extent during the refinement, we combine the encoder of one model with the decoder of the other model, i.e., the encoder of the \textit{Koopman-SI} model with the decoder of the \textit{Koopman-RL} model and vice versa. As can be seen from the lower part of Table \ref{tab:koopman_embedding_analysis}, the combinations of encoders and decoders from different models still yield low reconstruction losses, suggesting that the Koopman models before and after the refinement operate with very similar lifted spaces. Interestingly, using the encoder of the \textit{Koopman-SI} model together with the decoder of the \textit{Koopman-RL} model produces almost identical results to using the full \textit{Koopman-SI} autoencoder. Conversely, using the \textit{Koopman-RL} encoder in conjunction with the \textit{Koopman-SI} decoder produces very similar results to using the full \textit{Koopman-RL} autoencoder. These observations indicate that the end-to-end refinement affected the encoder to a larger extent than the decoder.

\subsection{eNMPC with adapted bounds}\label{sec:results_CS2_2}
    One key advantage of MPC controllers over model-free policies is that they may be able to adapt to changing control settings, such as shifted bounds or a change in the objective function. Model-free policies cannot adapt to such changes, as constraints and objective function are learned implicitly by the policy through the reward signal of the environment. Therefore, any change to the environment would require a retraining of the policy.
    
\begin{figure*}[ht]
    	\centering
    	\includegraphics[width=0.7\paperwidth]{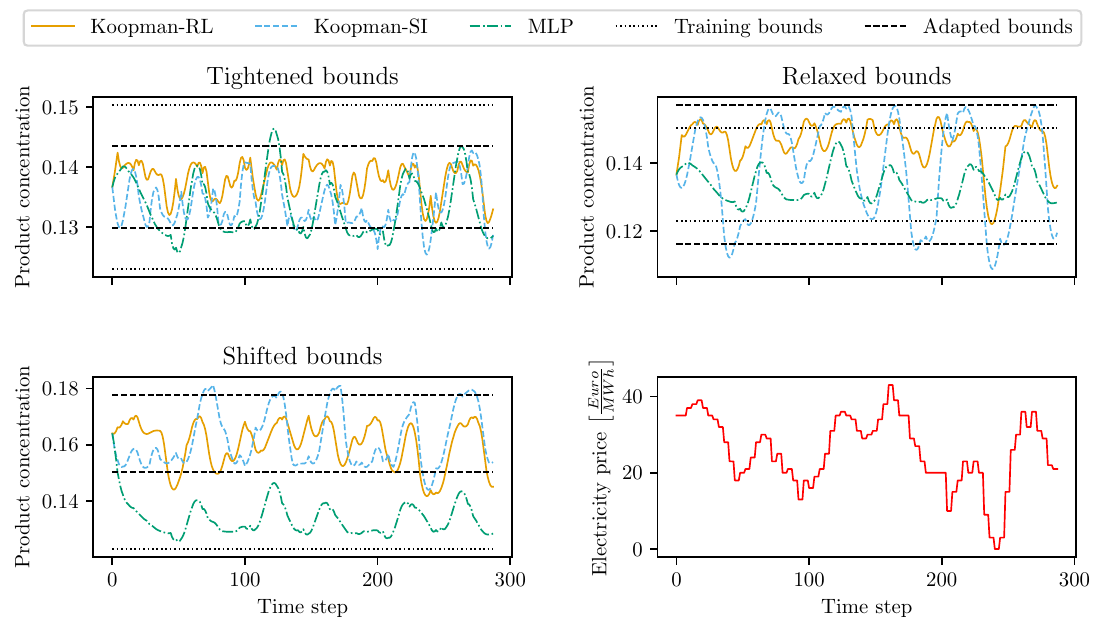}
    	\caption{eNMPC with adapted $c$ bounds: comparison of the controller behavior, given a randomly sampled continuous electricity price trajectory. The original bounds used during RL training are shown in dotted black lines, whereas the shifted bounds are shown in dashed black lines. Best viewed in color.}
    	\label{fig:trajectory_comparison_CS2_2}
    \end{figure*}

    We shift the bounds of the product concentration variable $c$ and rerun the test (cf. Subsection \ref{sec:results_CS2}) without retraining. We examine three different constraint adaptions: relaxed bounds, tightened bounds, and a shifted feasible region of $c$, i.e., a different target product. Table \ref{tab:cs2_shifted_bounds} shows the adapted bounds of $c$ in all three cases and reports the frequency of constraint violations that result from applying the controllers. Additionally, Figure \ref{fig:trajectory_comparison_CS2_2} illustrates the resulting trajectories of $c$ when using the controllers on a three-day test episode for the different bound adaptations. We waive reporting the production costs in Table \ref{tab:cs2_shifted_bounds} as they become meaningless in the presence of vast differences in the rates of constraint violations.
    
    \begin{table*}[ht]
    \centering
    \setlength{\extrarowheight}{0.05cm}
    \caption{Adapted lower (lb) and upper (ub) bounds of $c$ and constraint violations resulting from applying the controllers without retraining. The constraint violation column reports the percentage of control steps that result in constraint violations.}
    \label{tab:cs2_shifted_bounds}
        \begin{tabular}{l|rr|rrr}
            \toprule
                      & \multicolumn{2}{c|}{Adapted bounds of $c$} & \multicolumn{3}{c}{Constr. viols. [\%]}  \\
                      & lb              & ub             & \textit{MLP}     & \textit{Koopman-SI}   & \textit{Koopman-RL}\\ \hline
            Tightened & 0.1299          & 0.1435         &  49.20           &  19.53                &  1.73              \\
            Relaxed   & 0.1162          & 0.1572         &   0.24           &   4.41                &  0.13              \\
            Shifted   & 0.1504          & 0.1777         & 100.00           &  14.94                &  5.83              \\
            \bottomrule
        \end{tabular}
    \end{table*}
    
    Both Table \ref{tab:cs2_shifted_bounds} and Figure \ref{fig:trajectory_comparison_CS2_2} show that the \textit{Koopman-RL} controller is best at adapting to a change in the control setting, even though RL trained it for optimal performance in a somewhat different setting. Even though the \textit{tightened bounds} and \textit{shifted bounds} cases lead to more frequent constraint violations than in the original setting (cf. Table \ref{tab:cs2_performance}), the \textit{Koopman-RL} controller apparently still profits from its end-to-end refinement, as constraint violations are markedly rarer than for the \textit{Koopman-SI} controller. In the \textit{tightened bounds} and \textit{relaxed bounds} cases, the relative difference in the frequency of constraint violations compared to the \textit{Koopman-SI} controller stays similar to the original case (cf. Table \ref{tab:cs2_performance}). In the \textit{shifted bounds} case, the \textit{Koopman-RL} controller produces more constraint violations than in the other two cases, which is expected, as the RL training discouraged ever entering the \textit{shifted bounds} range.
    
    Unsurprisingly, the \textit{MLP} controller is agnostic to a change in the control setting and fails in the \textit{tightened bounds} and \textit{shifted bounds} cases. In the \textit{relaxed bounds} case, it causes few constraint violations due to its unaltered behavior. Again, all constraint violations are violations of the non-negativity constraint of the product storage, which we did not relax. As the \textit{MLP} controller respects the original training bounds of the state variables $c$ and $T$, it does not take advantage of the enhanced flexibility potential of the process in the \textit{relaxed bounds} case.

\section{Conclusion}\label{sec:conclusion} 
    This article presents a method for end-to-end RL of Koopman models for optimal performance in (e)NMPC applications. We exploit the modeling approach by \cite{korda2018linear} to construct Koopman-MPC controllers with convex underlying OCPs, and \textit{cvxpylayers} (\cite{Agrawal2019differentiable}) for post-optimal sensitivity analysis, thus turning the Koopman-MPC controllers into automatically differentiable policies that can be trained using state-of-the-art RL methods like PPO (\cite{schulman2017proximal}). Applying a trained Koopman-MPC controller has a comparatively low computational cost due to the convexity of the OCPs. Therefore, our method should be able to produce predictive controllers that are real-time capable for potentially large control systems.

    We demonstrate the effectiveness of our method by comparing the control performance to those of MPC utilizing models trained by system identification and model-free neural network controllers trained by RL. Specifically, we investigate two control applications derived from a CSTR model: (i) a setpoint tracking problem (NMPC, Section \ref{sec:results_CS1}), and (ii) a demand response problem with hard constraints on state variables (eNMPC, Section \ref{sec:results_CS2}). Our RL-trained Koopman models consistently outperform their counterparts trained by system identification.

    Additionally, we alter the feasible region of a state variable in eNMPC and test the controllers without retraining (Section \ref{sec:results_CS2_2}). Such changes in the control setting present a significant challenge to data-driven control. Due to the explicit constraints in the Koopman-eNMPCs, end-to-end training using RL preserves the controllers inherent adaptability to changes in the environment. Thus, even though the Koopman-eNMPCs lose some of the gains in performance from the RL refinement procedure, they can adapt better to the changes to the environment than the Koopman-eNMPCs trained by system identification alone. As expected, the neural network controllers cannot adapt as they implicitly learn the constraints of their environment through the rewards received during training.

    We validated our method on nonlinear but relatively small case studies, far from the complexity of many real-world systems. Future work should therefore test the approach on more complex systems and, eventually, real-world systems. However, applying our method to more complex control problems would likely necessitate training for more iterations. Additionally, it would increase the computational burden associated with each iteration due to larger Koopman surrogate models and more expensive environment steps. Both factors would increase the overall computational burden of the training process and could render our approach computationally intractable for very large problems. Future work should therefore explore ways to increase the data efficiency and stability of the learning process. To this end, differentiable Koopman (e)NMPC could be combined with model-based RL algorithms (\cite{Janner2019MBPO}) or policy optimization algorithms that exploit the differentiability of environments that are based on mechanistic simulation models (\cite{xu2022accelerated, mayfrank2024task}). Another avenue of possible future research is combining RL and differentiable Koopman (e)NMPC to learn task-optimal disturbance estimators (\cite{SonKwonJProc2022}) instead of refining the dynamic model using RL.

\section*{Declaration of Competing Interest}
We have no conflict of interest.

\section*{Acknowledgements}
This work was performed as part of the Helmholtz School for Data Science in Life, Earth and Energy (HDS-LEE) and received funding from the Helmholtz Association of German Research Centres.

We thank the anonymous reviewer for raising the idea of using differentiable Koopman (e)NMPC to learn task-optimal disturbance estimators for offset-free MPC (see Conclusion).
We also want to thank our former ICE-1 colleagues Danimir T. Doncevic and Florian J. Baader for fruitful discussions and literature recommendations.

\section*{Declaration of generative AI and AI-assisted technologies in the writing process}
During the preparation of this work Daniel Mayfrank used ChatGPT and Grammarly in order to correct grammar and spelling and to improve style of writing. After using these tools, all authors reviewed and edited the content as needed and take full responsibility for the content of the publication.

\section*{Nomenclature}

\subsection*{Abbreviations}
\noindent\begin{table}[H]
\begin{tabular}{ll}
    CSTR      & continuous stirred-tank reactor \\
    (e)(N)MPC & (economic) (nonlinear) model \\
              & predictive control \\
    HVAC      & heating, ventilation and air \\
              & conditioning \\
    KKT       & Karush-Kuhn-Tucker \\
    LICQ      & linear independence constraint \\
              & qualification \\
    MDP       & Markov decision process \\
    MLP       & multilayer perceptron \\
    MSE       & mean squared error \\
    OCP       & optimal control problem \\
    PPO       & proximal policy optimization \\
    RL        & reinforcement learning \\
    SCS       & strict complementary slackness \\
    SI        & system identification \\
    SOSC      & second-order sufficient condition
\end{tabular}
\end{table}

\subsection*{Greek Symbols}
\begin{table}[H]
\begin{tabular}{ll}
    $\alpha$        & learning rate \\
    $\beta$         & reward calculation hyperparameter \\
    $\gamma$        & reward discount factor \\
    $\epsilon$      & clipping hyperparameter \\
    $\bm{\theta}$   & learnable parameters of actor/predictor \\
    $\lambda$       & generalized advantage estimation \\
                    & hyperparameter \\
    $\bm{\mu}$      & expected value for action selection \\
    $\bm{\pi}$      & policy \\
    $\rho$          & CSTR production rate \\
    $\bm{\sigma}$   & standard deviation for action selection \\
    $\bm{\phi}$     & learnable parameters of critic \\
    $\Phi$          & MPC stage cost \\
    $\bm{\psi}$     & encoder MLP
\end{tabular}
\end{table}

\subsection*{Latin Symbols}
\begin{table}[H]
\begin{tabular}{ll}
    $\bm{a}$        & solution point of an optimization \\
                    & problem \\
    $\bm{A}$        & autoregressive part of Koopman \\
                    & dynamics matrix \\
    $\bm{B}$        & external input part of Koopman \\
                    & dynamics matrix \\
    $c$             & CSTR product concentration \\
    $\bm{C}$        & decoder matrix of Koopman model \\
    $F$             & CSTR coolant flow rate \\
    $\bm{g}$        & inequality constraints \\
    $K_{\text{PPO}}$& number of epochs per update \\
    $M$             & penalty factor for slack variables \\
    $M_{\text{PPO}}$& minibatch size \\
    $\bm{\mathcal{N}}$ & normal distribution \\
    $N_{\text{PPO}}$& number of parallel actors \\
    $p$             & electricity price \\
    $r$             & reward \\
    $\bm{s}$        & slack variables \\
    $t$             & indicates time-dependency of variable \\
    $T$             & CSTR temperature \\
    $\mathrm{T}$    & set of discrete time steps \\
    $T_{\text{PPO}}$& control steps between updates to \\
                    & actor and critic \\
    $\bm{u}$        & control variables \\
    $\bm{x}$        & system state variables \\
    $\bm{z}$        & Koopman state variables

\end{tabular}
\end{table}

\subsection*{Subscripts}
\begin{table}[H]
\begin{tabular}{ll}
    $\text{ext}$& indicates external input \\
    $\text{lb}$ & lower bound \\
    $\text{ss}$ & steady-state \\
    $t$         & discrete time step \\
    $\text{ub}$ & upper bound
\end{tabular}
\end{table}

\subsection*{Superscripts}
\begin{table}[H]
\begin{tabular}{ll}
    $*$         & indicates optimality
\end{tabular}
\end{table}

\section*{Author contributions}
Daniel Mayfrank: Conceptualization, Methodology, Software, Investigation, Writing - original draft, Visualization

\noindent Alexander Mitsos: Conceptualization, Writing - review \& editing, Supervision, Funding acquisition

\noindent Manuel Dahmen: Conceptualization, Methodology, Writing - review \& editing, Supervision, Funding acquisition

\appendix

\bibliographystyle{apalike}
  \renewcommand{\refname}{Bibliography}

  \bibliography{extracted.bib}

\begin{thebibliography}{}

\bibitem[Abadi et~al., 2016]{abadi2016tensorflow}
Abadi, M., Barham, P., Chen, J., Chen, Z., Davis, A., Dean, J., Devin, M., Ghemawat, S., Irving, G., Isard, M., et~al. (2016).
\newblock Tensor{F}low: a system for large-scale machine learning.
\newblock In {\em 12th USENIX Symposium on Operating Systems Design and Implementation (OSDI 16)}, pages 265--283.

\bibitem[Agrawal et~al., 2019a]{Agrawal2019differentiable}
Agrawal, A., Amos, B., Barratt, S., Boyd, S., Diamond, S., and Kolter, J.~Z. (2019a).
\newblock Differentiable convex optimization layers.
\newblock {\em Advances in Neural Information Processing Systems}, 32:9558--9570.

\bibitem[Agrawal et~al., 2019b]{agrawal2019differentiating}
Agrawal, A., Barratt, S., Boyd, S., Busseti, E., and Moursi, W.~M. (2019b).
\newblock Differentiating through a cone program.
\newblock {\em arXiv preprint arXiv:1904.09043}.

\bibitem[Albalawi and Hameed, 2023]{Albalawi2023EconomicKoopman}
Albalawi, F. and Hameed, S.~W. (2023).
\newblock Koopman-based economic model predictive control for nonlinear systems.
\newblock In {\em 2023 9th International Conference on Control, Decision and Information Technologies (CoDIT)}, pages 822--828.

\bibitem[Amos et~al., 2018]{amos2018differentiable}
Amos, B., Jimenez, I., Sacks, J., Boots, B., and Kolter, J.~Z. (2018).
\newblock Differentiable {MPC} for end-to-end planning and control.
\newblock {\em Advances in Neural Information Processing Systems}, 31:8299--8310.

\bibitem[Amos and Kolter, 2017]{amos2017optnet}
Amos, B. and Kolter, J.~Z. (2017).
\newblock Optnet: Differentiable optimization as a layer in neural networks.
\newblock In {\em International Conference on Machine Learning}, pages 136--145.

\bibitem[Angeli et~al., 2011]{angeli2011average}
Angeli, D., Amrit, R., and Rawlings, J.~B. (2011).
\newblock On average performance and stability of economic model predictive control.
\newblock {\em IEEE Transactions on Automatic Control}, 57(7):1615--1626.

\bibitem[Arbabi et~al., 2018]{arbabi2018datadrivenkoopman}
Arbabi, H., Korda, M., and Mezic, I. (2018).
\newblock A data-driven {K}oopman model predictive control framework for nonlinear flows.
\newblock {\em arXiv preprint arXiv:1804.05291}.

\bibitem[Beal et~al., 2018]{beal2018gekko}
Beal, L., Hill, D., Martin, R., and Hedengren, J. (2018).
\newblock Gekko optimization suite.
\newblock {\em Processes}, 6(8):106.

\bibitem[Bengio et~al., 2015]{Bengio2015_curriculum}
Bengio, S., Vinyals, O., Jaitly, N., and Shazeer, N. (2015).
\newblock Scheduled sampling for sequence prediction with recurrent neural networks.
\newblock {\em Advances in Neural Information Processing Systems}, 28:1171--1179.

\bibitem[Boyd et~al., 2004]{boyd2004convex}
Boyd, S., Boyd, S.~P., and Vandenberghe, L. (2004).
\newblock {\em Convex {O}ptimization}.
\newblock Cambridge University Press.

\bibitem[Bruder et~al., 2019]{bruder2019modelingcontrolsoftrobots}
Bruder, D., Gillespie, B., Remy, C.~D., and Vasudevan, R. (2019).
\newblock Modeling and control of soft robots using the {K}oopman operator and model predictive control.
\newblock {\em arXiv preprint arXiv:1902.02827}.

\bibitem[Brunton and Kutz, 2022]{brunton2022data}
Brunton, S.~L. and Kutz, J.~N. (2022).
\newblock {\em Data-driven Science and Engineering: Machine Learning, Dynamical Systems, and Control}.
\newblock Cambridge University Press.

\bibitem[Chen et~al., 2019]{chen2019gnu}
Chen, B., Cai, Z., and Berg{\'e}s, M. (2019).
\newblock Gnu-{RL}: A precocial reinforcement learning solution for building {HVAC} control using a differentiable {MPC} policy.
\newblock In {\em Proceedings of the 6th ACM International Conference on Systems for Energy-Efficient Buildings, Cities, and Transportation}, pages 316--325.

\bibitem[Cibulka et~al., 2020]{Cibulka20204}
Cibulka, V., Haniš, T., Korda, M., and Hromčík, M. (2020).
\newblock Model predictive control of a vehicle using {K}oopman operator.
\newblock {\em IFAC-PapersOnLine}, 53(2):4228--4233.
\newblock 21st IFAC World Congress.

\bibitem[Domke, 2012]{domke2012generic}
Domke, J. (2012).
\newblock Generic methods for optimization-based modeling.
\newblock In {\em Artificial Intelligence and Statistics}, pages 318--326.

\bibitem[Dormand and Prince, 1980]{dormand1980family}
Dormand, J.~R. and Prince, P.~J. (1980).
\newblock A family of embedded {R}unge-{K}utta formulae.
\newblock {\em Journal of Computational and Applied Mathematics}, 6(1):19--26.

\bibitem[Du et~al., 2015]{du2015time}
Du, J., Park, J., Harjunkoski, I., and Baldea, M. (2015).
\newblock A time scale-bridging approach for integrating production scheduling and process control.
\newblock {\em Computers \& Chemical Engineering}, 79:59--69.

\bibitem[Engstrom et~al., 2020]{engstrom2020implementation}
Engstrom, L., Ilyas, A., Santurkar, S., Tsipras, D., Janoos, F., Rudolph, L., and Madry, A. (2020).
\newblock Implementation matters in deep policy gradients: A case study on {PPO} and {TRPO}.
\newblock {\em arXiv preprint arXiv:2005.12729}.

\bibitem[Fiacco and Ishizuka, 1990]{fiacco1990sensitivity}
Fiacco, A.~V. and Ishizuka, Y. (1990).
\newblock Sensitivity and stability analysis for nonlinear programming.
\newblock {\em Annals of Operations Research}, 27(1):215--235.

\bibitem[Fiacco and Kyparisis, 1985]{fiacco1985sensitivity}
Fiacco, A.~V. and Kyparisis, J. (1985).
\newblock Sensitivity analysis in nonlinear programming under second order assumptions.
\newblock In {\em Systems and {O}ptimization}, pages 74--97. Springer.

\bibitem[Flores-Tlacuahuac and Grossmann, 2006]{flores2006simultaneous}
Flores-Tlacuahuac, A. and Grossmann, I.~E. (2006).
\newblock Simultaneous cyclic scheduling and control of a multiproduct {CSTR}.
\newblock {\em Industrial \& Engineering Chemistry Research}, 45(20):6698--6712.

\bibitem[Folkestad and Burdick, 2021]{Folkestad2021}
Folkestad, C. and Burdick, J.~W. (2021).
\newblock Koopman {NMPC}: Koopman-based learning and nonlinear model predictive control of control-affine systems.
\newblock In {\em 2021 IEEE International Conference on Robotics and Automation (ICRA)}, pages 7350--7356.

\bibitem[Fujimoto et~al., 2018]{fujimoto2018addressing}
Fujimoto, S., Hoof, H., and Meger, D. (2018).
\newblock Addressing function approximation error in actor-critic methods.
\newblock In {\em International Conference on Machine Learning}, pages 1587--1596.

\bibitem[Goodfellow et~al., 2016]{Goodfellow-et-al-2016}
Goodfellow, I., Bengio, Y., and Courville, A. (2016).
\newblock {\em Deep Learning}.
\newblock MIT Press.

\bibitem[Gros and Zanon, 2019]{gros2019data}
Gros, S. and Zanon, M. (2019).
\newblock Data-driven economic {NMPC} using reinforcement learning.
\newblock {\em IEEE Transactions on Automatic Control}, 65(2):636--648.

\bibitem[Haarnoja et~al., 2018]{haarnoja2018soft}
Haarnoja, T., Zhou, A., Abbeel, P., and Levine, S. (2018).
\newblock Soft actor-critic: Off-policy maximum entropy deep reinforcement learning with a stochastic actor.
\newblock In {\em International Conference on Machine Learning}, pages 1861--1870.

\bibitem[Hussein et~al., 2017]{hussein2017imitation}
Hussein, A., Gaber, M.~M., Elyan, E., and Jayne, C. (2017).
\newblock Imitation learning: A survey of learning methods.
\newblock {\em ACM Computing Surveys (CSUR)}, 50(2):1--35.

\bibitem[Iwata and Kawahara, 2022]{iwata2022data}
Iwata, T. and Kawahara, Y. (2022).
\newblock Data-driven end-to-end learning of pole placement control for nonlinear dynamics via {K}oopman invariant subspaces.
\newblock {\em arXiv preprint arXiv:2208.08883}.

\bibitem[Janner et~al., 2019]{Janner2019MBPO}
Janner, M., Fu, J., Zhang, M., and Levine, S. (2019).
\newblock When to trust your model: Model-based policy optimization.
\newblock {\em Advances in Neural Information Processing Systems}, 32:12498--12509.

\bibitem[Kalman et~al., 1960]{kalman1960contributions}
Kalman, R.~E. et~al. (1960).
\newblock Contributions to the theory of optimal control.
\newblock {\em Boletín de la Sociedad Matemática Mexicana}, 5(2):102--119.

\bibitem[Karush, 1939]{karush1939minima}
Karush, W. (1939).
\newblock Minima of functions of several variables with inequalities as side constraints.
\newblock {\em M. Sc. Dissertation. Department of Mathematics, University of Chicago}.

\bibitem[Kingma and Ba, 2014]{kingma2014adam}
Kingma, D.~P. and Ba, J. (2014).
\newblock Adam: A method for stochastic optimization.
\newblock {\em arXiv preprint arXiv:1412.6980}.

\bibitem[Koopman, 1931]{koopman1931hamiltonian}
Koopman, B.~O. (1931).
\newblock Hamiltonian systems and transformation in {H}ilbert space.
\newblock {\em Proceedings of the National Academy of Sciences}, 17(5):315--318.

\bibitem[Korda and Mezi{\'c}, 2018]{korda2018linear}
Korda, M. and Mezi{\'c}, I. (2018).
\newblock Linear predictors for nonlinear dynamical systems: {K}oopman operator meets model predictive control.
\newblock {\em Automatica}, 93:149--160.

\bibitem[Lewis and Vrabie, 2009]{lewis2009reinforcement}
Lewis, F.~L. and Vrabie, D. (2009).
\newblock Reinforcement learning and adaptive dynamic programming for feedback control.
\newblock {\em IEEE Circuits and Systems Magazine}, 9(3):32--50.

\bibitem[Lewis et~al., 2012]{lewis2012reinforcement}
Lewis, F.~L., Vrabie, D., and Vamvoudakis, K.~G. (2012).
\newblock Reinforcement learning and feedback control: Using natural decision methods to design optimal adaptive controllers.
\newblock {\em IEEE Control Systems Magazine}, 32(6):76--105.

\bibitem[Lusch et~al., 2018]{lusch2018deep}
Lusch, B., Kutz, J.~N., and Brunton, S.~L. (2018).
\newblock Deep learning for universal linear embeddings of nonlinear dynamics.
\newblock {\em Nature Communications}, 9(1):1--10.

\bibitem[Mayfrank et~al., 2024]{mayfrank2024task}
Mayfrank, D., Ahn, N.~Y., Mitsos, A., and Dahmen, M. (2024).
\newblock Task-optimal data-driven surrogate models for {eNMPC} via differentiable simulation and optimization.
\newblock {\em arXiv preprint arXiv:2403.14425}.

\bibitem[McBride and Sundmacher, 2019]{mcbride2019overview}
McBride, K. and Sundmacher, K. (2019).
\newblock Overview of surrogate modeling in chemical process engineering.
\newblock {\em Chemie Ingenieur Technik}, 91(3):228--239.

\bibitem[Mnih et~al., 2013]{mnih2013playing}
Mnih, V., Kavukcuoglu, K., Silver, D., Graves, A., Antonoglou, I., Wierstra, D., and Riedmiller, M. (2013).
\newblock Playing {A}tari with deep reinforcement learning.
\newblock {\em arXiv preprint arXiv:1312.5602}.

\bibitem[Narasingam and Kwon, 2019]{Nasasingam2019}
Narasingam, A. and Kwon, J. S.-I. (2019).
\newblock Koopman {L}yapunov-based model predictive control of nonlinear chemical process systems.
\newblock {\em AIChE Journal}, 65(11):e16743.

\bibitem[Narasingam and Kwon, 2020]{Nasasingam2020}
Narasingam, A. and Kwon, J. S.-I. (2020).
\newblock Application of {K}oopman operator for model-based control of fracture propagation and proppant transport in hydraulic fracturing operation.
\newblock {\em Journal of Process Control}, 91:25--36.

\bibitem[Nocedal and Wright, 1999]{nocedal1999numerical}
Nocedal, J. and Wright, S.~J. (1999).
\newblock {\em Numerical {O}ptimization}.
\newblock Springer.

\bibitem[{Open Power System Data}, 2020]{open_power_system_data_2020}
{Open Power System Data} (2020).
\newblock Open power system data.
\newblock \url{https://data.open-power-system-data.org/time_series/} (accessed on {2022-08-29)}.

\bibitem[Paszke et~al., 2019]{paszke2017automatic}
Paszke, A., Gross, S., Massa, F., Lerer, A., Bradbury, J., Chanan, G., Killeen, T., Lin, Z., Gimelshein, N., Antiga, L., et~al. (2019).
\newblock Py{T}orch: An imperative style, high-performance deep learning library.
\newblock {\em Advances in Neural Information Processing Systems}, 32:8024--8035.

\bibitem[Pistikopoulos et~al., 2020]{pistikopoulos2020multi}
Pistikopoulos, E.~N., Diangelakis, N.~A., and Oberdieck, R. (2020).
\newblock {\em Multi-{P}arametric {O}ptimization and {C}ontrol}.
\newblock John Wiley \& Sons.

\bibitem[Proctor et~al., 2018]{proctor2018generalizing}
Proctor, J.~L., Brunton, S.~L., and Kutz, J.~N. (2018).
\newblock Generalizing {K}oopman theory to allow for inputs and control.
\newblock {\em SIAM Journal on Applied Dynamical Systems}, 17(1):909--930.

\bibitem[Quinonero-Candela et~al., 2008]{quinonero2008dataset}
Quinonero-Candela, J., Sugiyama, M., Schwaighofer, A., and Lawrence, N.~D. (2008).
\newblock {\em Dataset {S}hift in {M}achine {L}earning}.
\newblock {MIT} Press.

\bibitem[Ralph and Dempe, 1995]{ralph1995directional}
Ralph, D. and Dempe, S. (1995).
\newblock Directional derivatives of the solution of a parametric nonlinear program.
\newblock {\em Mathematical Programming}, 70(1):159--172.

\bibitem[Schulman et~al., 2015a]{schulman2015trust}
Schulman, J., Levine, S., Abbeel, P., Jordan, M., and Moritz, P. (2015a).
\newblock Trust region policy optimization.
\newblock In {\em International Conference on Machine Learning}, pages 1889--1897.

\bibitem[Schulman et~al., 2015b]{schulman2015high}
Schulman, J., Moritz, P., Levine, S., Jordan, M., and Abbeel, P. (2015b).
\newblock High-dimensional continuous control using generalized advantage estimation.
\newblock {\em arXiv preprint arXiv:1506.02438}.

\bibitem[Schulman et~al., 2017]{schulman2017proximal}
Schulman, J., Wolski, F., Dhariwal, P., Radford, A., and Klimov, O. (2017).
\newblock Proximal policy optimization algorithms.
\newblock {\em arXiv preprint arXiv:1707.06347}.

\bibitem[Silver et~al., 2014]{silver2014deterministic}
Silver, D., Lever, G., Heess, N., Degris, T., Wierstra, D., and Riedmiller, M. (2014).
\newblock Deterministic policy gradient algorithms.
\newblock In {\em International Conference on Machine Learning}, pages 387--395.

\bibitem[Son et~al., 2021]{SonKwon2021}
Son, S.~H., Choi, H.-K., and Kwon, J. S.-I. (2021).
\newblock Application of offset-free {K}oopman-based model predictive control to a batch pulp digester.
\newblock {\em AIChE Journal}, 67(9):e17301.

\bibitem[Son et~al., 2022a]{SonKwon2022}
Son, S.~H., Choi, H.-K., Moon, J., and Kwon, J. S.-I. (2022a).
\newblock Hybrid {K}oopman model predictive control of nonlinear systems using multiple {EDMD} models: An application to a batch pulp digester with feed fluctuation.
\newblock {\em Control Engineering Practice}, 118:104956.

\bibitem[Son et~al., 2022b]{SonKwonJProc2022}
Son, S.~H., Narasingam, A., and Kwon, J. S.-I. (2022b).
\newblock Development of offset-free {K}oopman {L}yapunov-based model predictive control and mathematical analysis for zero steady-state offset condition considering influence of {L}yapunov constraints on equilibrium point.
\newblock {\em Journal of Process Control}, 118:26--36.

\bibitem[Sutton and Barto, 2018]{sutton2018reinforcement}
Sutton, R.~S. and Barto, A.~G. (2018).
\newblock {\em Reinforcement {L}earning: An {I}ntroduction}.
\newblock MIT press.

\bibitem[Virtanen et~al., 2020]{virtanen2020scipy}
Virtanen, P., Gommers, R., Oliphant, T.~E., Haberland, M., Reddy, T., Cournapeau, D., Burovski, E., Peterson, P., Weckesser, W., Bright, J., et~al. (2020).
\newblock Sci{P}y 1.0: fundamental algorithms for scientific computing in {P}ython.
\newblock {\em Nature Methods}, 17(3):261--272.

\bibitem[Williams et~al., 2016]{williams2016extending}
Williams, M.~O., Hemati, M.~S., Dawson, S.~T., Kevrekidis, I.~G., and Rowley, C.~W. (2016).
\newblock Extending data-driven {K}oopman analysis to actuated systems.
\newblock {\em IFAC-PapersOnLine}, 49(18):704--709.

\bibitem[Xu et~al., 2022]{xu2022accelerated}
Xu, J., Makoviychuk, V., Narang, Y., Ramos, F., Matusik, W., Garg, A., and Macklin, M. (2022).
\newblock Accelerated policy learning with parallel differentiable simulation.
\newblock {\em arXiv preprint arXiv:2204.07137}.

\bibitem[Yao et~al., 2022]{yao2022wild}
Yao, H., Choi, C., Cao, B., Lee, Y., Koh, P. W.~W., and Finn, C. (2022).
\newblock Wild-time: A benchmark of in-the-wild distribution shift over time.
\newblock {\em Advances in Neural Information Processing Systems}, 35:10309--10324.

\bibitem[Yin et~al., 2022]{yin2022embedding}
Yin, H., Welle, M.~C., and Kragic, D. (2022).
\newblock Embedding {K}oopman optimal control in robot policy learning.
\newblock In {\em 2022 IEEE/RSJ International Conference on Intelligent Robots and Systems (IROS)}, pages 13392--13399. IEEE.

\bibitem[Zaremba and Sutskever, 2014]{zaremba2014learning}
Zaremba, W. and Sutskever, I. (2014).
\newblock Learning to execute.
\newblock {\em arXiv preprint arXiv:1410.4615}.

\bibitem[Zhang et~al., 2021]{zhang2021adaptive}
Zhang, M., Marklund, H., Dhawan, N., Gupta, A., Levine, S., and Finn, C. (2021).
\newblock Adaptive risk minimization: Learning to adapt to domain shift.
\newblock {\em Advances in Neural Information Processing Systems}, 34:23664--23678.

\end{thebibliography}


\begin{thebibliography}{}

\bibitem[Domahidi et~al., 2013]{domahidi2013ecos}
Domahidi, A., Chu, E., and Boyd, S. (2013).
\newblock {ECOS}: An {SOCP} solver for embedded systems.
\newblock In {\em 2013 European control conference (ECC)}, pages 3071--3076. IEEE.

\bibitem[O'Donoghue et~al., 2016]{SCS_solver}
O'Donoghue, B., Chu, E., Parikh, N., and Boyd, S. (2016).
\newblock Conic optimization via operator splitting and homogeneous self-dual embedding.
\newblock {\em Journal of Optimization Theory and Applications}, 169(3):1042--1068.

\bibitem[Schulman et~al., 2017]{schulman2017proximal}
Schulman, J., Wolski, F., Dhariwal, P., Radford, A., and Klimov, O. (2017).
\newblock Proximal policy optimization algorithms.
\newblock {\em arXiv preprint arXiv:1707.06347}.

\end{thebibliography}

\end{document}


\ifx\REVIEW\undefined
\begin{@twocolumnfalse}
\fi

  \thispagestyle{firststyle}

  \begin{center}
    \begin{large}
      \textbf{\mytitle}
    \end{large} \\
    \myauthor
  \end{center}

  \vspace{0.5cm}

  \begin{footnotesize}
    \affil
  \end{footnotesize}

  \vspace{0.5cm}
  
\ifx\REVIEW\undefined
\end{@twocolumnfalse}
\fi

\begin{center}
\setlength{\extrarowheight}{0.05cm}
    \captionof{table}{Hyperparameters in NMPC. Where possible, the notation is consistent with the PPO paper (\cite{schulman2017proximal}).}
    \begin{tabular}{llll}
        \toprule
                                            & Hyperparameter  & Value                     & Description \\ \hline
        \multicolumn{2}{l}{\textbf{General}}                  &                           &  \\
                                            & $\bm{\sigma}$   & $(0.05, 0.05)^\intercal$  & standard deviation for action selection \\
                                            & $\gamma$        & 0.95                      & reward discount factor \\
                                            & $\lambda$       & 0.95                      & generalized advantage estimation hyperparameter \\
                                            & $\epsilon$      & 0.2                       & clipping hyperparameter \\
                                            & $N_\text{PPO}$  & 1                         & number of parallel actors \\
                                            & $T_\text{PPO}$  & 2,048                     & control steps between updates to actor and critic \\
                                            & $M_\text{PPO}$  & 64                        & minibatch size \\
                                            & optimizer       & Adam                      & optimizer used for updates to actor and critic \\
                                            & $\alpha_\text{actor}$&  $10^{-4}$           & learning rate of actor \\
                                            & $\alpha_\text{critic}$& $3\cdot10^{-4}$     & learning rate of critic  \\
                                            & episodes        & 5,000                     & number of episodes per training run \\
                                            &                 &                           &  \\
        \multicolumn{2}{l}{\textbf{Koopman MPC policies}}     &                           &    \\
                                            & $K_\text{PPO}$  & 5                         & number of epochs per update \\
                                            & max. gradient norm & 100.0                  & gradient clipping value for actor update \\
                                            & solver          & SCS                       & solver for Koopman OCPs \\
                                            &                 &                           & (\cite{SCS_solver})  \\
                                            & max. iters.     & 500                       & maximum number of iterations in Koopman OCP \\
                                            &                 &                           & solver  \\
                                            &                 &                           &    \\
        \multicolumn{2}{l}{\textbf{MLP policies}}             &                           &    \\
                                            & $K_\text{PPO}$  & 10                        & number of epochs per update  \\
                                            & max. gradient norm & 40.0                   & gradient clipping value for actor update \\
        \bottomrule
    \end{tabular}
\end{center}
\clearpage

\begin{center}
\setlength{\extrarowheight}{0.05cm}
    \captionof{table}{Hyperparameters in eNMPC. Where possible, the notation is consistent with the PPO paper (\cite{schulman2017proximal}).}
    \begin{tabular}{llll}
        \toprule
                                            & Hyperparameter  & Value                     & Description \\ \hline
        \multicolumn{2}{l}{\textbf{General}}                  &                           &             \\
                                            & $\beta$         & $5\cdot10^{-5}$           & reward calculation hyperparameter \\
                                            & $\bm{\sigma}$   & $(0.05, 0.05)^\intercal$  & standard deviation for action selection\\
                                            & $\gamma$        & 0.95                      & reward discount factor    \\
                                            & $\lambda$       & 0.95                      & generalized advantage estimation hyperparameter\\
                                            & $\epsilon$      & 0.2                       & clipping hyperparameter  \\
                                            & $N_\text{PPO}$  & 1                         & number of parallel actors \\
                                            & $T_\text{PPO}$  & 2,048                     & control steps between updates to actor and critic\\
                                            & $M_\text{PPO}$  & 64                        & minibatch size            \\
                                            & optimizer       & Adam                      & optimizer used for updates to actor and critic \\
                                            &                 &                           &             \\
        \multicolumn{2}{l}{\textbf{Koopman MPC policies}}     &                           &    \\
                                            & episodes        & 5,000                     & number of episodes per training run \\
                                            & $K_\text{PPO}$  & 5                         & number of epochs per update   \\
                                            & $\alpha_\text{actor}$&  $10^{-5}$           & learning rate of actor   \\
                                            & $\alpha_\text{critic}$& $2\cdot10^{-5}$     & learning rate of critic    \\
                                            & max. gradient norm & 100.0                  & gradient clipping value for actor update \\
                                            & solver          & ECOS                      & solver for Koopman OCPs\\
                                            &                 &                           & (\cite{domahidi2013ecos}) \\
                                            & max. iters.     & 500                       & maximum number of iterations in Koopman OCP \\
                                            &                 &                           & solver  \\
                                            & $M$             & 10,000                    & penalty factor for slack variable usage\\
                                            &                 &                           &             \\
        \multicolumn{2}{l}{\textbf{MLP policies}}             &                           &    \\
                                            & episodes        & 20,000                    & number of episodes per training run \\
                                            & $K_\text{PPO}$  & 10                        & number of epochs per update \\
                                            & max. gradient norm & 40.0                   & gradient clipping value for actor update\\
                                            & $\alpha_\text{actor}$&  $10^{-4}$           & learning rate of actor \\
                                            & $\alpha_\text{critic}$& $3\cdot10^{-4}$     & learning rate of critic \\
        \bottomrule
    \end{tabular}
\end{center}

\clearpage
  \appendix

\bibliographystyle{apalike}
  \renewcommand{\refname}{Bibliography}

  \bibliography{extracted.bib}